\PassOptionsToPackage{table}{xcolor}
\documentclass[letterpaper]{article} 

\usepackage{times}  
\usepackage{helvet}  
\usepackage{courier}  
\usepackage{url}  
\usepackage{graphicx}  
\usepackage{wrapfig}

\usepackage[utf8]{inputenc} 
\usepackage[T1]{fontenc}    
\usepackage{booktabs}       
\usepackage{amsfonts}       
\usepackage{nicefrac}       
\usepackage{microtype}      
\usepackage{tikz}
\usetikzlibrary{shapes.misc, positioning}
\usepackage{thmtools,thm-restate}
\usepackage{parskip}
\usepackage{authblk}
\usepackage{amsmath}  
\usepackage{mathtools}
\usepackage{amsthm}

\usepackage{amsmath,amsfonts,bm}

\def\eqref#1{equation~\ref{#1}}

\def\1{\bm{1}}
\def\Y{\mathcal{Y}}

\DeclareMathAlphabet{\mathsfit}{\encodingdefault}{\sfdefault}{m}{sl}
\SetMathAlphabet{\mathsfit}{bold}{\encodingdefault}{\sfdefault}{bx}{n}

\newcommand{\E}{\mathbb{E}}

\newcommand{\R}{\mathbb{R}}

\DeclareMathOperator*{\argmax}{arg\,max}
\DeclareMathOperator*{\argmin}{arg\,min}

\usepackage{optidef}
\usepackage{algorithm}
\usepackage{algpseudocode}

\theoremstyle{plain}
\newtheorem{theorem}{Theorem}[section]
\newtheorem{proposition}[theorem]{Proposition}
\newtheorem{lemma}[theorem]{Lemma}
\newtheorem{corollary}[theorem]{Corollary}
\theoremstyle{definition}
\newtheorem{definition}[theorem]{Definition}

\theoremstyle{remark}

\usepackage[table]{xcolor}

\newcommand{\set}[1]{\mathcal{#1}}
\usepackage[numbers,sort]{natbib}

\usepackage{basic}

\newcommand{\loki}{\textsc{Loki}}

\title{Geometry-Aware Adaptation for Pretrained Models}

 \author[$\dagger$]{Nicholas Roberts}
 \author[$\dagger$]{Xintong Li}
 \author[$\dagger$]{Dyah Adila}
 \author[$\dagger$]{Sonia Cromp}
 \author[$\dagger$]{\\Tzu-Heng Huang}
 \author[$\dagger$]{Jitian Zhao}
\author[$\dagger$]{Frederic Sala}

 \affil[$\dagger$]{University of Wisconsin-Madison}
 \affil[ ]{\footnotesize{\texttt{\{nick11roberts, fredsala\}@cs.wisc.edu}}}
 \affil[ ]{\footnotesize{\texttt{\{xli2224, adila, cromp, thuang273, jzhao326\}@wisc.edu}}}

\begin{document}

\maketitle

\begin{abstract}

Machine learning models---including prominent zero-shot models---are often trained on datasets whose labels are only a small proportion of a larger label space.
Such spaces are commonly equipped with a metric that relates the labels via distances between them.
We propose a simple approach to exploit this information to adapt the trained model to reliably predict new classes---or, in the case of zero-shot prediction, to improve its performance---without any additional training.
Our technique is a drop-in replacement of the standard prediction rule, swapping $\argmax$ with the Fréchet mean.
We provide a comprehensive theoretical analysis for this approach, studying 
(i) finite-sample bounds separating probability-estimation error, irreducible prediction risk, and the effect of observing only a subset of labels, 
(ii) {conditions on the observed label set under which every label can be a Fr\'echet minimizer in several metric spaces}, and
(iii) {a polynomial-time procedure for choosing the next class to maximize the locus}.
Empirically, using easily-available external metrics, our proposed approach, $\loki$, gains up to 29.7\% relative improvement over SimCLR on ImageNet and scales to hundreds of thousands of classes.
When no such metric is available, $\loki$ can use self-derived metrics from class embeddings and obtains a 10.5\% improvement on pretrained zero-shot models such as CLIP. 

\end{abstract}


\section{Introduction}


The use of pretrained models is perhaps the most significant recent shift in machine learning.
Such models can be used out-of-the-box, either to predict classes observed during pretraining (as in ImageNet-trained ResNets \cite{resnet}) or to perform zero-shot classification on any set of classes \cite{clip}.
While this is exciting, label spaces are often so huge that models are unlikely to have seen \emph{even a single point} with a particular label. 
Without additional modification, pretrained models will simply fail to reliably predict such classes.
Instead, users turn to fine-tuning---which requires additional labeled data and training cycles and so sacrifices much of the promise of zero-shot usage. 

%
%
%

How can we adapt pretrained models to predict new classes, without fine-tuning or retraining?
At first glance, this is challenging: predictive signal comes from labeled training data.
However, \emph{relational} information between the classes can be exploited to enable predicting a class even when there are no examples with this label in the training set.
Such relational data is commonly available, or can be constructed with the help of knowledge bases, ontologies, or powerful foundation models \cite{stewart2017label}.

%
%
%
%

How to best exploit relational structure remains unclear, with a number of key challenges: 
We might wish to know what particular subset of classes is rich enough to enable predicting many (or all) remaining labels.
This is crucial in determining whether a training set is usable or, even with the aid of structure, insufficient.
It is also unclear how approaches that use relational information interact with the statistical properties of learning, such as training sample complexity.
Finally, performing adaptation requires an efficient and scalable algorithm. 

\begin{figure*}[t!]
\centerline{
\includegraphics[width=0.8\columnwidth]{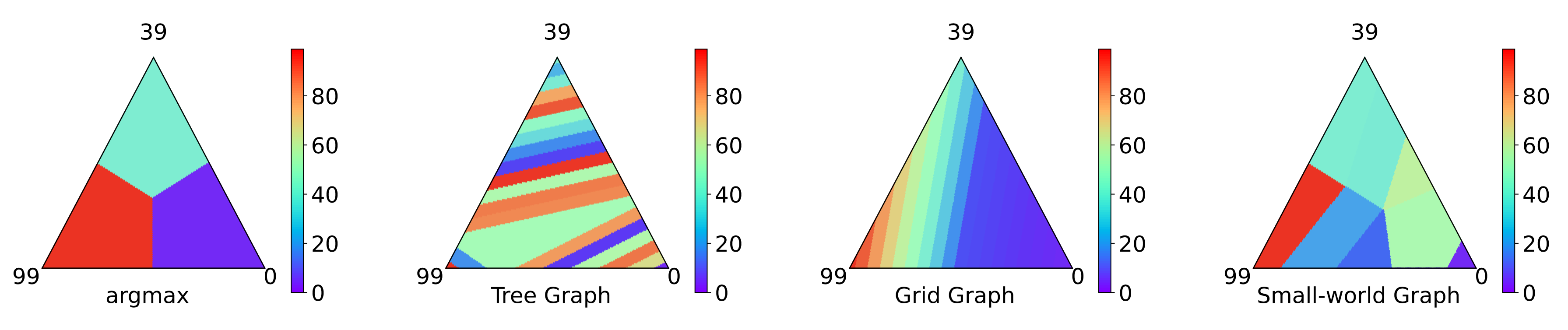}
}
\caption{Classification regions in the probability simplex of 3-class classifiers faced with a 100-class problem. 
The probability simplex using $\argmax$ prediction can only output one of three classes. 
$\loki$ uses the entire probability vector to navigate the class metric space, leading to more prediction regions. 
(Left) regions from $\argmax$ prediction. 
(Centers, Right) classification regions from $\loki$.  }
\label{fig:banner}
\end{figure*}

This work addresses these challenges.
It proposes a simple and practical approach to learning in structured label spaces, with theoretical guarantees.
First, we offer a simple way to translate the soft outputs (i.e., probability vectors) produced by \emph{any} supervised learning model into a more general model that can exploit geometric information for label structure.
In other words, our approach, called $\loki$,\footnote{Refers to the `locus' (plural: \textit{loci}) of the Fréchet mean.} is a simple \emph{adaptor} for pretrained models. 
$\loki$ can be applied via a fixed linear transformation of the model outputs.
$\loki$'s simplicity makes it applicable to a broad range of settings while enabling very high-cardinality predictions subject to a potentially small model output budget---we provide a visualization of this key idea in Figure~\ref{fig:banner}. 

Theoretically, we provide a rich set of results for the metric-based adaptation setting.
{First, we bound the error in the Fr\'echet objective---the probability-weighted sum of squared distances from a candidate prediction to the labels---caused by probability estimation. Using this, we give a finite-sample excess-risk guarantee for squared distance in any finite label metric, under a supervised Gaussian feature model. The bounds account for label-metric diameter, feature dimension, per-class sample allocation, and a separate term for restricting the observed label set.} 
{Next, we characterize minimum and identifying locus covers for several metric structures.}
Finally we show how to exploit this result in an active learning-like approach to selecting points that will improve deficient training datasets.

Experimentally, we show that using structural information improves prediction in high-cardinality settings. 
We demonstrate the strength of the active learning-based approach to dataset expansion over random baselines.
Finally, and most excitingly, we show that even in zero-shot models like CLIP, where it is possible to produce probability vectors over any possible class, the use of our adaptor leads to a \textbf{19.53\%} relative improvement. 

\section{Background}
First, we introduce the problem setting, notation, and mathematical tools that we will use. 
Afterward, we discuss how $\loki$ relates to prior work. 

\paragraph{Problem Setting} As in conventional supervised learning, we have a dataset $(x_1, y_1), \ldots, (x_n, y_n)$ drawn from a distribution on $\mathcal{X} \times \Y$, where $\mathcal{X}$ and $\mathcal{Y}$ are the input and label spaces. In our setting, $N := |\mathcal{Y}|$ is finite but large; often $N \gg n$---so that many labels will simply not be found in the training dataset. We let $\Lambda \subseteq \set{Y}$ with $|\Lambda| = K$ be the set of observed labels.
For convenience of notation, we also define $\boldsymbol{\lambda} := [\lambda_i]_{i=1}^K$, {$\boldsymbol{y} := [y_j]_{j=1}^N$} to be the vectors of elements in $\Lambda$ and $\mathcal{Y}$.

In addition to our dataset, we have access to a relational structure on $\Y$. We assume that $\Y$ is a metric space with metric (distance) $d : \Y^2 \rightarrow \mathbb{R}$; $d$ encodes the relational structure of the label space. 
Specifically, we model this metric space using a graph, $G = (\set{Y}, \set{E})$ where $\set{E} \subseteq \set{Y}^2 \times \mathbb{R}_{+}$ is a set of edges relating the labels and the standard shortest-path distance $d: \set{Y}^2 \to \mathbb{R}_{\geq 0}$. 
In addition to its use in prediction, the metric $d$ can be used for evaluating a model by measuring, for example, {$\frac{1}{n} \sum_{i=1}^n d^2(f(x_i), y_i)$}---the analogue of the empirical square loss.

\paragraph{Fréchet Mean Estimator} 
Drawing on ideas from structured prediction \cite{Ciliberto2016, Rudi18}, we use a simple predictor that exploits the label metric. It relies on computing the \textit{Fréchet mean} \cite{Frechet1948}, given by
\begingroup \begin{align} m_{\boldsymbol{\lambda}}(\bm{w}) := \argmin_{y \in \set{Y}} \sum_{i=1}^K \bm{w}_i d^2(y, \lambda_i),
\label{def:fm}
\end{align}\endgroup
{The weights are nonnegative and not all zero. Normalizing them to sum to one leaves the minimizers unchanged, so we use probability weights in the closed simplex $\Delta^{K-1}$.} 
The Fr\'{e}chet mean generalizes the barycenter to metric spaces and is often used in geometric machine learning \cite{Lou20}. 

\paragraph{Locus of the Fréchet mean.} 
The \textit{locus of the Fréchet mean} is the set of all Fréchet means under different weights \cite{pca_phylogenetic}. 
%
{We write it as $\Pi(\boldsymbol{\lambda}) := \bigcup_{\bm{w} \in \Delta^{K-1}} m_{\boldsymbol{\lambda}}(\bm{w})$.}

{The locus contains every label that is a Fr\'echet minimizer for some probability weights on $\Lambda$. A fixed tie-breaking rule or a particular trained classifier may produce only a subset of these labels.} 
Intuitively, we can think of the locus for a given dataset as describing how usable it is for predicting beyond the observed labels. 
{Trivially, if $\Lambda=\set{Y}$, then $\Pi(\boldsymbol{\lambda})=\set{Y}$.}
{We are interested in the case in which $\Lambda\subset\set{Y}$ yet $\Pi(\boldsymbol{\lambda})=\set{Y}$, or at least $|\Pi(\boldsymbol{\lambda})|$ is sufficiently large.} 

\subsection{Relation to Prior work}
$\loki$ is primarily related to two areas: zero-shot learning and structured prediction. 

\paragraph{Zero-Shot Learning}
Like zero-shot learning (ZSL), $\loki$ is capable of predicting unobserved classes.
Our framework is most closely related to \textit{generalized} ZSL, which uses side information to predict both observed and unobserved classes. 
Many types of external knowledge are used in ZSL, including text \cite{text_zsl, text_zsl_2, text_zsl_3}, attributes \cite{attr_zsl_1, attr_zsl_2, attr_zsl_3}, knowledge graphs \cite{KG_zsl_1, KG_zsl_2, KG_zsl_3}, ontologies \cite{ont_zsl_1}, and logical rules \cite{ont_zsl_2}. 
Our work is most closely related to ZSL approaches that rely on knowledge graph information. 
Often, ZSL methods that use knowledge graph information rely on the use of graph neural network architectures \cite{KG_zsl_1, Kampffmeyer2018RethinkingKG, kipf2017semisupervised}.
\textit{However, we note that these architectures can be heavyweight and can be challenging to scale up to extremely large graphs, whereas $\loki$ does not have architectural requirements and scales linearly in the size of the graph when $K$ is small.} 



\paragraph{Structured Prediction}
Structured prediction (SP) operates in label spaces that are endowed with algebraic or geometric structure \cite{bakir_predicting_2007, NIPS2015_52d2752b}. 
This includes problems such as predicting permutations \cite{NEURIPS2018_b3dd760e}, non-Euclidean and manifold regression \cite{Petersen2016FrchetRF, NEURIPS2018_f6185f0e}, and learning on graphs \cite{NEURIPS2019_ea697987}. 
$\loki$ is the most immediately related to the latter, however, any finite metric space can be represented as a graph, which further lends to the flexibility of our approach. 
Even in discrete structured prediction settings, the cardinality of the label space may be combinatorially large. 
\textit{As such, $\loki$ can be viewed as a simple method for adapting classifiers to structured prediction. }

The Fréchet mean has been used 
in structured prediction---but in approaches requiring training. 
%
In \cite{Ciliberto2016}, 
$ \hat{f}(x) =$  $\arg\min_{y \in \Y}$  $\frac{1}{n}\sum_{i=1}^n \alpha_i(x)d^2(y, \bm{y}_i)$, 
where $\alpha(x) = (K + n \nu I)^{-1} K_x$. 
$K$ is the kernel matrix for a kernel 
$k: \mathcal{X} \times \mathcal{X} \rightarrow \mathbb{R}$, so that $K_{i,j} = k(x_i, x_j)$, $(K_x)_i = k(x, x_i)$. 
$\nu$ is a regularization parameter. 
In other words, the weight $w_i$ corresponding to $\bm{y}_i$ is the average produced by solving a kernel regression problem at all points $x_k$ in the dataset where $y_k = \bm{y}_i$. 
It has also used in weak supervision (WS) for metric-equipped label spaces
\cite{vishwakarma2022lifting, shin2022universalizing}, where the goal is to produce labeled data for training structured prediction models. 



\section{Framework}
We introduce our framework---$\loki$. 
We show how $\loki$ can be used to adapt any supervised classifier over a set of $K$ classes to a much richer set of possible class predictions. 
{$\loki$ uses the classifier's probabilities as weights in the Fr\'echet objective and selects a minimizer, which may be an unobserved class. The rule takes probabilities as input. Note that logits can be converted using softmax.} 
Next, we show how $\loki$ can be expressed as a \textit{fixed} linear transformation of a model's outputs. 
Finally, we show that $\loki$ relates to standard classification.

\subsection{$\loki$: Adapting Pretrained Models}
\label{sec:framework}
We describe our approach to adapting pretrained classifiers--trained on a set of classes $\Lambda$---to the metric geometry of the label space $\mathcal{Y}$, enabling the prediction of unobserved classes. 

We model unobserved classes $\mathcal{Y} \setminus \Lambda$ using the Fréchet mean among observed classes weighted by their \textit{prediction probabilities} $P(y=\lambda_i| x \text{ and } y \in \Lambda)$. 
We denote the vector of model outputs as $\mathbf{P}_{y|x} := [\mathbf{P}_{\lambda_i|x}]_{i=1}^K = [P(y=\lambda_i| x \text{ and } y \in \Lambda)]_{i=1}^K$. 
Then predictions using $\loki$ are given by
$$ \hat{y} \in m_{\boldsymbol{\lambda}}(\mathbf{P}_{y|x}) = \argmin_{y \in \set{Y}} \sum_{i=1}^K \mathbf{P}_{\lambda_i|x} d^2(y, \lambda_i). $$

\subsection{$\loki$ as a linear transformation of model outputs}
Most standard classifiers output a vector of prediction probabilities, $\mathbf{P}_{y|x}$, whose entries correspond to the confidence of predicting a specific class. 
Predictions are typically given by $\hat{y} \in \argmax_{i \in [K]} (\mathbf{P}_{y|x})_i$. 
$\loki$ generalizes this prediction rule when viewed as a linear transformation of $\mathbf{P}_{y|x}$. 
Consider the $\loki$ prediction rule
    $\hat{y} \in m_{\boldsymbol{\lambda}}(\mathbf{P}_{y|x}) = \argmin_{y \in \set{Y}} \sum_{i=1}^K \mathbf{P}_{\lambda_i|x} d^2(y, \lambda_i) = \argmax_{j \in [N]} (\mathbf{D} \mathbf{P}_{y|x})_j,$
where $\mathbf{D}_{j, i} = \left[ -d^2(y_j, \lambda_i) \right]$; $\mathbf{D} \in \R^{N \times K}$ is the matrix of negative squared distances between the observed classes and the rest of the label space. 
Thus $\loki$ can be used within standard classification pipelines when the model output $\mathbf{P}_{y|x}$ is multiplied by the fixed matrix $\mathbf{D}$.


\subsection{Generalizing Standard Classification}
We provide a simple intuition for our approach.
The fact that $\loki$ reduces to standard classification among the observed classes has several implications. 
This includes the idea that under our framework, forms of few-shot, zero-shot, hierarchical, and partial label learning all reduce to standard classification when additional metric information is introduced. 

\paragraph{Generalizing the arg max prediction rule} In the absence of this metric information---a situation that we model using the complete graph and setting $\Lambda = \mathcal{Y}$---our framework also recovers standard classification. 
Indeed, both in terms of error modeling and in terms of inter-class similarity, the intuition of standard classification and of the 0-1 loss are captured well by the unweighted complete graph---simply treat all differences equally. 
This graph is given as $K_{N} := (\set{Y}, \set{Y}^2 \times \{1\})$---i.e., every label is equidistant from every other label. 
Plugging this into $\loki$, we obtain the following:
\begin{align*}
    \hat{y} \in m_{\boldsymbol{\lambda}}(\mathbf{P}_{y|x}) &= \argmin_{y \in \set{Y}} \sum_{i=1}^K \mathbf{P}_{\lambda_i|x} d^2(y, \lambda_i) 
    = \argmin_{y \in \set{Y}} \sum_{i=1}^K \mathbf{P}_{\lambda_i|x} \boldsymbol{1}\{ y \neq \lambda_i \} 
    = \argmax_{i \in [K]} \mathbf{P}_{\lambda_i|x}, 
\end{align*}
which is exactly the standard classification prediction rule.

\paragraph{Generalizing the 0-1 loss via the expected squared distance} \emph{The expected squared distance $\E[d^2(y, \hat{y})]$ is the standard loss function for many structured prediction problems}. Note that \textbf{accuracy fails in such settings}---since it cannot distinguish between small and large errors. This is most clearly seen in the extreme case of regression, where test accuracy will virtually always be zero no matter how good a trained model is. At the other extreme, the complete graph, this loss function becomes the standard 0-1 loss: $\E[d^2(y, \hat{y})] = \E[\boldsymbol{1}\{y \neq \hat{y}\}]$. 
As an adaptor for structured label spaces, we use the empirical version of this loss to evaluate $\loki$. 
Note that the expected squared distance subsumes other metrics as well. 
For example, when $\set{Y} = \mathbb{R}$, we can derive the standard MSE by setting $d(y, \hat{y}) = |y - \hat{y}|$, which is just the standard L1 distance metric. 
Other scores such as recall at top-k can be similarly obtained at the cost of $d(\cdot, \cdot)$ being a true metric. 
In other words, $\E[d^2(y, \hat{y})]$ is an very general metric that supports any metric space, and we use it throughout this work. 


\section{Theoretical Results}
%
%
%

\paragraph{Challenges and Opportunities}
The $\argmax$ of per-class model probabilities is a ubiquitous component of classification pipelines in machine learning. 
In order to predict unobserved classes using metric space information, $\loki$ replaces this standard component. 
As a simple but significant change to standard pipelines, $\loki$ opens up a new area for fundamental questions. 
There are three main flavors of theoretical questions that arise in this setting:
\begin{enumerate}
    \item How does the performance of $\loki$ change as a function of the number of samples? 
    \item What minimal sets of observed classes are required to predict any class in the metric space? 
    \item How can we acquire new classes that maximize the total number of classes we can predict? 
\end{enumerate}
We analyze these questions for several metric spaces used in practice. First, we give a perturbation bound for the Fr\'echet objective on any finite metric space and finite-sample bounds for squared label-metric loss under an identity-covariance Gaussian feature model. 
Second, we characterize the sets of observed classes that are required to enable prediction of any class, and show how this set differs for various types of metric spaces of interest. 
Finally, we provide an active learning algorithm for selecting additional classes to observe so as to maximize the number of classes that can be predicted and we characterize the types of metric spaces for which the locus can be computed efficiently. 
These results provide a strong theoretical grounding for $\loki$. 




\begingroup
\subsection{Sample Complexity}
How does the number of training examples affect the squared-distance error of $\loki$ on observed and unobserved classes? We estimate class means, localize the Gaussian probability weights near observed means, and use the covering geometry of the radius-$R$ feature ball. The resulting bounds involve the radius, input dimension, observed-class count, sample size, and confidence level explicitly.

Let $\mathcal Y$ be a finite label space with metric $d(\cdot,\cdot)$ and diameter $D$. Each label $y$ has a feature mean $\mu_y\in B_2(0,R)\subset\mathbb R^d$, where $R\ge0$ and scalar $d\ge1$ denotes input dimension. Let $2\le K\le|\mathcal Y|$, and choose the observed classes $\Lambda=\{y_i\}_{i\in[K]}\subseteq\mathcal Y$ so that
\[
\max_{y\in\mathcal Y}\min_{i\in[K]}\|\mu_y-\mu_{y_i}\|_2
\le\frac{2R}{K^{1/d}-1}.
\]
Such a subset of exactly $K$ labels exists by a covering argument (we include it in the appendix). The $n$ training examples are independent and identically distributed, with labels uniform on $\Lambda$; independent test examples have labels uniform on $\mathcal Y$. In both distributions, features conditional on class $y$ have distribution $\mathcal N(\mu_y,I_d)$.
Use the per-class median-of-means estimator (specified in the appendix), and form $\hat{\mathbb P}(y_i\mid x)$ with known identity covariance and equal priors. Define
\[
\Psi_{\mathbf w}(y)=\sum_{i\in[K]}\mathbf w_i d^2(y,y_i),\qquad
\hat y_*\in m_\Lambda([\hat{\mathbb P}(y_i\mid x)]_{i\in[K]})
=\arg\min_{y\in\mathcal Y}\sum_{i\in[K]}\hat{\mathbb P}(y_i\mid x)d^2(y,y_i),
\]
with fixed tie breaking. Write $\mathbf P(x)=[\mathbb P(y_i\mid x)]_{i\in[K]}$ for the exact posterior conditioned on label membership in $\Lambda$, and let
\[
\mathcal C_\Lambda=4\mathbb E_{(x,y_*)\sim\mathcal D_{\mathrm{test}}}
\Psi_{\mathbf P(x)}(y_*).
\]

\begin{theorem}[$\loki$ sample complexity]
\label{theorem:learning}
For $0<\delta<1$ and $n\ge24K\log(4dK/\delta)$, with probability at least $1-\delta$ over training,
\[
\mathbb E_{(x,y_*)\sim\mathcal D_{\mathrm{test}}}[d^2(y_*,\hat y_*)]
\le\mathcal C_\Lambda+
O\!\left(D^2\left(\sqrt d+\frac{2R}{K^{1/d}-1}+\sqrt{2\log K}\right)
\sqrt{\frac{dK\log(4dK/\delta)}{n}}\right).
\]
If there is an $L>0$ such that $d(y,z)\le L\|\mu_y-\mu_z\|_2$ for all labels, the baseline and estimation term can both be made explicit:
\begin{align*}
\mathbb E[d^2(y_*,\hat y_*)]
&\le4L^2\left(2\sqrt d+\frac{2R}{K^{1/d}-1}+\sqrt{2\log K}\right)^2\\
&\quad+O\!\left(L^2\left(2\sqrt d+\frac{2R}{K^{1/d}-1}+\sqrt{2\log K}\right)
\sqrt{\frac{dK\log(4dK/\delta)}{n}}\right).
\end{align*}
All fitted-predictor expectations are conditional on training, and the hidden constants are universal. Both bounds may be capped at $D^2$.
\end{theorem}

For Euclidean distances, with $\mu_y=y$, set $L=1$. The second bound gives a baseline for noise and missing labels, plus an estimation term that decays as $n^{-1/2}$. 
For graph distances, we use the first bound.  
Note that the baseline $\mathcal C_\Lambda$ need not equal Bayes risk or the risk with exact probabilities. 
\par\endgroup


\subsection{Optimal Label Subspaces}
\label{sec:optimal_subspaces}
Next we characterize the subset of distinct labels required to predict any label using our label model with respect to various types of metric spaces. 

We first consider label spaces whose metric is a tree graph---such metrics are, for example, related to performing hierarchical classification and weak supervision (where only partial labels are available). 
We consider a special type of tree called a phylogenetic tree, in which only the leaves can be designated as labels---phylogenetic trees are commonly used to relate the labels of image classification datasets. 
Afterwards we perform a similar analysis for grid graphs, which are important for label spaces that encode spatial information. 
Finally, we discuss the case in which no useful metric information is available, i.e., the complete graph.

Our goal in this section is to characterize the properties and size of $\{\lambda_i\}_{i=1}^K$ in each of these metric spaces such that we still have $\Pi({\bf\Lambda}) = \set{Y}$. 
We characterize `optimal' subsets of classes in each of the spaces under a certain notions of optimality.  
We provide several relevant definitions pertaining to this concept, starting with a notion of being able to predict any possible class using observed classes. 

\begin{definition}[Locus cover]
\label{def:loc_cover}
Given a set $\Lambda \subseteq \set{Y}$ for which we construct a tuple of its elements $\boldsymbol{\Lambda}$, if it holds that $\Pi(\boldsymbol{\Lambda}) = \set{Y}$, then $\Lambda$ is a locus cover. 
\end{definition}

Definition~\ref{def:loc_cover} captures the main idea of $\loki$---using some set of observed classes for which we can train classifiers, we would like to be able to predict additional unobserved classes using the geometry that relates the observed and unobserved classes. {Here ``reachable'' means a minimizer for some choice of probability weights.} 
We refine this Definition to describe the trivial case that defaults to standard classification and the nontrivial case for which $\loki$ moves beyond standard classification. 

\begin{definition}[Trivial locus cover]
\label{def:trivial_loc_cover}
If $\Lambda = \set{Y}$, then $\Lambda$ is the trivial locus cover. 
\end{definition}

This Definition captures the notion of observing all of the classes in the label space. 
Here, all of the elements of $\set{Y}$ are trivially reachable using $\loki$. 

\begin{definition}[Nontrivial locus cover]
\label{def:nontrivial_loc_cover}
A locus cover $\Lambda$ is nontrivial if $\Lambda \neq \set{Y}$. 
\end{definition}

$\loki$ is more useful and interesting when faced with a nontrivial locus cover---under Definition~\ref{def:nontrivial_loc_cover}, {some proper subset $\Lambda$ suffices to include every label in $\set Y$ among the possible minimizers.} 

\begin{definition}[Minimum locus cover]
\label{def:min_loc_cover}
{A locus cover is minimum if its cardinality is no greater than that of any other locus cover.} 
\end{definition}

{When the label space is large, we would like to use the smallest observed set $\Lambda$ whose locus contains every label in $\set{Y}$.} 
{We characterize minimum locus covers for trees in which every vertex is a label and for full rectangular grids.} 
It is worth noting that the minimum locus cover need not be unique for a fixed graph. 

\begin{definition}[Identifying locus cover]
\label{def:identifying_loc_cover}
Given a set $\Lambda \subseteq \set{Y}$, if $\Lambda$ is a locus cover where $\forall\, y \in \set{Y}, \, \exists\, \mathbf{w} \in \Delta^{|\Lambda|-1}$ such that $m_{\boldsymbol{\Lambda}}(\mathbf{w}) = \{ y \}$,
then $\Lambda$ is an identifying locus cover. 
\end{definition}

The Fréchet mean need not be unique---as an $\argmin$, it returns a set of minimizers. 
In certain metric spaces, the minimum locus cover can yield large sets of minimizers---this is undesirable, as it makes predicting a single class challenging. 
{Definition~\ref{def:identifying_loc_cover} requires each label to be the unique Fr\'echet minimizer for some choice of weights; that choice may depend on the label. This does not require uniqueness for every weight vector.} 

\begingroup
An identifying cover makes every label attainable independently of the tie-breaking rule, when weights can range freely over the simplex. A particular classifier need not produce all of these weights. Ordinary covers need not have this property: on the $3\times3$ grid, weights on opposite corners give an objective depending only on the coordinate sum. The set-valued locus has nine vertices, but a fixed ordering can select at most one vertex from each of its five coordinate-sum layers.
\par\endgroup

\begingroup
\paragraph{Complexity in general metric spaces.}
We can construct minimum covers for several of the metric spaces below. In general, however, finding a minimum cover is NP-hard.
\begin{theorem}[Complexity of minimum locus covers]
\label{thm:locus_cover_hardness}
Let $\mathcal Y$ be a nonempty finite metric space with rational distances, given by its distance matrix. Finding a smallest locus cover or a smallest identifying locus cover is NP-hard. 
For either type of cover, deciding whether a cover of size at most $k$ exists is NP-complete, where $k$ is given in binary. These results hold even when all nonzero distances are $1$ or $2$.
\end{theorem}
We prove both results by reducing Vertex Cover to locus-cover selection (Appendix~\ref{app:locus_cover_hardness}). We also show that linear programming can check whether a proposed set is a cover in polynomial time, giving membership in NP. As in Definition~\ref{def:min_loc_cover}, a minimum cover is one with the fewest labels. Our construction uses a finite metric, which can be represented as a complete weighted graph. The identifying-cover result does not establish hardness for unweighted graph distances.

\par\endgroup

\begin{definition}[Pairwise decomposable]
\label{def:pairwise_decomposable}
Given $\Lambda \subseteq \set{Y}$, $\Pi(\Lambda)$ is called pairwise decomposable when it holds that $\Pi(\Lambda) = \cup_{\lambda_1, \lambda_2 \in \Lambda} \Pi(\{\lambda_1, \lambda_2\}).$ 
\end{definition}

In many cases, the locus can be written in a more convenient form---the union of the locus of pairs of nodes. 
We refer to this definition as pairwise decomposability. 
Later, we shall see that pairwise decomposability is useful in computing the locus in polynomial time.

\paragraph{Trees}
Many label spaces are endowed with a tree metric in practice: hierarchical classification, in which the label space includes both classes and superclasses, partial labeling problems in which internal nodes can represent the prediction of a set of classes, and the approximation of complex or intractable metrics using a minimum spanning tree. 
We show that for our purposes, trees have certain desirable properties that make them easy to use with $\loki$---namely that we can easily identify a locus cover that satisfies both Definition~\ref{def:min_loc_cover} and Definition~\ref{def:identifying_loc_cover}. 
Conveniently, we also show that any locus in any tree satisfies Definition~\ref{def:pairwise_decomposable}. 

We first note that the leaves of any tree yield the minimum locus cover. 
{Every vertex can be the unique Fr\'echet minimizer for some probability weights on the leaves.} 
This can be especially useful if the tree has long branches and few leaves. 
Additionally, for tree metric spaces, the minimum locus cover (Definition~\ref{def:min_loc_cover}) is also an identifying locus cover (Definition~\ref{def:identifying_loc_cover}). 
This follows from the construction of the weights in the proof of Theorem~\ref{thm:min_locus_trees} (shown in the Appendix) and the property that all paths in trees are unique. 
Finally, we note that any locus in any tree is pairwise decomposable---the proof of this is given in the Appendix (Lemma~\ref{lemma:tree_pairwise}). 
We will see later that this property yields an efficient algorithm for computing the locus. 

\paragraph{Phylogenetic Trees}
Image classification datasets often have a hierarchical tree structure, where only the leaves are actual classes, and internal nodes are designated as superclasses---examples include the ImageNet \cite{imagenet} and CIFAR-100 datasets \cite{cifar100}. 
Tree graphs in which only the leaf nodes are labeled are referred to as phylogenetic trees \cite{BILLERA2001733}. 
Often, these graphs are weighted, but unless otherwise mentioned, we assume that the graph is unweighted. 

For any arbitrary tree $T=(\set V, \set E)$, the set of labels induced by phylogenetic tree graph is $\set{Y} = \text{Leaves}(T)$. 
We provide a heuristic algorithm for obtaining locus covers for arbitrary phylogenetic trees in Algorithm~\ref{alg:phylo} (see Appendix). 
We prioritize adding endpoints of long paths to $\Lambda$, and continue adding nodes in this way until $\Pi(\Lambda)=\set{Y}$. 
\begingroup
{Phylogenetic loci need not be pairwise decomposable; Proposition~\ref{prop:phylo_nonpairwise} gives an unweighted counterexample. Algorithm~\ref{alg:phylo} returns a locus cover because it checks the locus exactly after each step and eventually includes every leaf if needed (Theorem~\ref{thm:algphylo}). For rational edge lengths, Algorithm~\ref{alg:locus_general} computes the leaf-only locus in polynomial time. Algorithm~\ref{alg:locus_pairwise} applies when pairwise decomposability has been established.}
\par\endgroup

\begingroup
\paragraph{Grid Graphs}
For spatial label spaces, consider a full rectangular Cartesian grid
$\mathcal Y=\prod_{j=1}^r\{0,\ldots,m_j\}$, {where $r$ is a positive integer and the $m_j$ are nonnegative integers}, with unit edges between vertices differing by one in a single coordinate. Its shortest-path metric is Manhattan distance. We exclude irregular grid subgraphs from the statements below.

When the grid has more than one vertex, any pair of opposite corners is a minimum locus cover (Theorem~\ref{thm:min_loc_grid}). This need not be an identifying cover: different vertices in the same distance layer can tie for every weighting of that pair. All corners together form an identifying locus cover, as proved in the same appendix. When at least two coordinate ranges are nonzero, this set has more than two elements and is therefore not a minimum ordinary cover; {for a path with at least two vertices}, its two endpoints are already a minimum identifying cover.

The locus of a pair is exactly the {coordinate box} between its endpoints (Lemma~\ref{lemma:locus_grid_subspace}). In two dimensions, this implies pairwise decomposability by Lemma~\ref{lemma:grid_pairwise}. The property fails in general in dimension three, as the counterexample following that lemma shows. Algorithm~\ref{alg:locus_general} applies in every dimension.
\par\endgroup

\paragraph{The Complete Graph}
The standard classification setting does not use relational information between classes. 
As before, we model this setting using the complete graph, and we show the expected result that in the absence of useful relational information, $\loki$ cannot help, and the problem once again becomes standard multiclass classification among observed classes. 
To do so, we show that {there is no nontrivial locus cover for the unweighted complete graph} (Theorem~\ref{thm:trivial_complete} in the Appendix). 



\subsection{Label Subspaces in Practice}
While it is desirable for the set of observed classes to form a minimum or identifying locus cover, it is often not possible to choose the initial set of observed classes a priori---these are often random. 
In this section, we describe the more realistic cases in which a random set of classes are observed and an active learning-based strategy to choose the next observed class. 
The aim of our active learning approach is, instead of randomly selecting the next observed class, to actively select the next class so as to maximize the total size of the locus---i.e., {the number of labels attainable as minimizers over all simplex weights.} 
Before maximizing the locus via active learning, we must first address a much more basic question: can we even efficiently compute the locus? 

\paragraph{Computing the Locus}
\begingroup
Given an observed set $\Lambda$, we can compute its locus in polynomial time for any finite metric with a rational distance matrix. For each label, Algorithm~\ref{alg:locus_general} uses linear programming to check whether some choice of weights makes that label a Fr\'echet minimizer. Each problem has $K$ variables and $O(N+K)$ constraints, where $K=|\Lambda|$ and $N=|\mathcal Y|$. We give the proof and runtime analysis in Appendix~\ref{app:general_locus_algorithm}. When the locus is pairwise decomposable, Algorithm~\ref{alg:locus_pairwise} provides another approach: compute the locus of each pair and take their union. Thus, computing the locus is polynomial-time, even though finding a minimum locus cover is NP-hard (Theorem~\ref{thm:locus_cover_hardness}).
\par\endgroup

\paragraph{Large Locus via Active Next-Class Selection}
\label{sec:active_selection}
We now turn to actively selecting the next class to observe in order to maximize the size of the locus. 
{For a finite metric given by a rational distance matrix, we can try each unobserved class and choose the one that gives the largest locus. For trees, we also give a direct rule based on the graph structure below.} 
Assume the initial set of $K$ observed classes are sampled at random from some distribution. 
We would like to actively select the $K+1$st class such that $|\Pi(\Lambda)|$ with $\Lambda = \{\lambda\}_{i=1}^{K+1}$ is as large as possible. 

\begingroup
\begin{corollary}[Optimal next-class selection]
\label{cor:general_next_class}
Let $\mathcal Y$ be a finite metric space given by a rational distance matrix, and let $\Lambda\subseteq\mathcal Y$ with $K<N$. We can find a class $v\in\mathcal Y\setminus\Lambda$ that maximizes $|\Pi(\Lambda\cup\{v\})|$ in polynomial time, using at most $N(N-K)$ linear feasibility problems.
\end{corollary}
\begin{proof}
For each of the $N-K$ unobserved classes, use Algorithm~\ref{alg:locus_general} to compute the locus after adding that class. Choose any class that gives the largest locus. Each candidate requires $N$ linear feasibility problems, each with at most $K+1$ variables and polynomial input bit length. The total runtime is therefore polynomial.
\end{proof}
\par\endgroup

\begingroup
\begin{theorem}[Optimal next-class selection on trees]
\label{thm:active_largest}
Let $T=(\mathcal Y,\mathcal E)$ be a finite tree with positive edge lengths, with every vertex a candidate label. Let $\emptyset\ne\Lambda\subseteq\mathcal Y$, with $\Lambda\ne\mathcal Y$, and let $H$ be the subtree induced by $\Pi(\Lambda)$.
If $\Pi(\Lambda)=\mathcal Y$, every $v\in\mathcal Y\setminus\Lambda$ is optimal and gives zero gain. Otherwise, each $v\notin\Pi(\Lambda)$ has a unique attachment vertex $b(v)\in H$, where its path first enters $H$, and
\[
|\Pi(\Lambda\cup\{v\})|-|\Pi(\Lambda)|
=|\Gamma(v,b(v))\setminus\{b(v)\}|.
\]
Consequently, choosing a vertex outside the locus that maximizes this path vertex count is optimal among all unobserved labels. For unit edge lengths, this is equivalent to maximizing $d(v,b(v))$.
\end{theorem}

The attachment paths and their vertex counts can be found by a tree traversal, so the rule is polynomial-time. With nonunit edge lengths one must maximize vertex count, not weighted distance. The result concerns full-vertex tree label spaces; it does not establish the same rule for phylogenetic trees whose candidate labels are only leaves. The general procedure in Corollary~\ref{cor:general_next_class} applies to those finite metrics as well.
\par\endgroup




\section{Experimental Results}
\label{sec:experiments}

\begin{table}[t!]
    \caption{
    \textbf{CIFAR-100.} Improving CLIP predictions using $\loki$. 
    Results are reported as $\E[d^2(y, \hat{y})]$ in the respective metric space. 
    CLIP-like zero-shot models can be improved using $\loki$ even without access to an external metric, and internal class embedding distances are used.  
    When an external metric is available, $\loki$ outperforms CLIP using the default CIFAR-100 hierarchy and WordNet. 
    } \label{table:clip_improver}
  \begin{center}
    \begin{tabular}{rr|ccc}
    \toprule 
    Model & Metric Space& $\argmax$ & $\loki$ & Relative Improvement \\
    \midrule
    \rowcolor[gray]{.9} 
    CLIP-RN50 \cite{clip}& Internal & 0.2922  & {\bf 0.2613}  & {\bf 10.57\%}  \\
    CLIP-ViT-L-14 \cite{clip}& Internal & 0.1588  & {\bf 0.1562}  &  {\bf1.63\%}  \\
    \rowcolor[gray]{.9} 
    ALIGN \cite{align}       & Internal & 0.1475  & {\bf 0.1430}  & {\bf 3.02\%}  \\
    \midrule
    CLIP-RN50 \cite{clip}& $K_{100}$           & 0.5941$^*$  & 0.5941$^*$  & 0.0\%$^*$  \\
    \rowcolor[gray]{.9} 
    CLIP-RN50 \cite{clip} & Default tree & 7.3528  & {\bf 7.1888} & {\bf 2.23\%} \\
    CLIP-RN50 \cite{clip} & WordNet tree & 24.3017  & {\bf 19.5549} & {\bf 19.53\%} \\
    \bottomrule
    \end{tabular}
    \\
    $^*$ methods equivalent under the complete graph as $\loki$ reduces to $\argmax$ prediction. \\
  \end{center}
\end{table}

%


In this section, we provide experimental results to validate the following claims:
\begin{enumerate}
    \item $\loki$ improves performance of zero-shot foundation models even with no external metric. 
    \item $\loki$ adapts to label spaces with a large number of unobserved classes. 
    \item {On synthetic trees, the active approach in Theorem~\ref{thm:active_largest} yields a larger locus than random class selection.}  
    \item {On ImageNet, the evaluated selection strategy yields lower squared-distance loss than random class selection.} 
    \item With $\loki$, calibration can improve \textit{accuracy}, even with no external metric. 
\end{enumerate}

\subsection{$\loki$ Improves Zero-Shot Models}
We evaluate the capability of $\loki$ to improve upon zero-shot models where all classes are observed.

\paragraph{Setup} 
Our experiment compares the zero-shot prediction performance of CLIP \cite{clip} on CIFAR-100 \cite{cifar100} {to predictions using $\loki$. The probability-weighted prediction rule requires converting logits to probabilities.} 
First, we consider the setting in which no external metric relating the labels is available, and instead derive internal metric information from Euclidean distances between text embeddings from the models using their respective text encoders. 
Second, we consider three external metric spaces for use with $\loki$: the complete graph $K_{100}$ and two phylogenetic trees: the default CIFAR-100 superclasses \cite{cifar100} and WordNet \cite{barz2019hierarchy}. 

\paragraph{Results} 
The results of this experiment are given in Table~\ref{table:clip_improver}. 
{Without an external metric, $\loki$ achieves lower squared-distance loss than the standard prediction rule in these experiments.} 
As expected, under the complete graph, our method becomes equivalent to the standard prediction mechanism used by CLIP. 
On the other hand, $\loki$ outperforms CLIP when using the default CIFAR-100 tree hierarchy and even more so when using the WordNet geometry, with a ${\bf 19.53\%}$ relative improvement in mean squared distance over the CLIP baseline.
We postulate that the strong performance using WordNet is due to the richer geometric structure compared to that of the default CIFAR-100 hierarchy.

\begin{wraptable}{R}{.35\linewidth}
\caption{
\textbf{PubMed.} $\loki$ improves baseline for all settings of $K$. 
The metric space is Euclidean distances applied to SimCSE embeddings. 
} \label{table:pubmed}
\begin{center}
    \begin{tabular}{r|cc}
    \toprule 
    $K$ & 5-NN  & 5-NN + \loki   \\
    \midrule
    \rowcolor[gray]{.9} 
    100 &1.68666   & {\bf 1.42591} \\
    250 &1.52374   & {\bf 1.47801} \\
    \rowcolor[gray]{.9} 
    500 &1.64074  & {\bf 1.45921} \\
    \bottomrule
    \end{tabular}
\end{center}
\caption{
\textbf{LSHTC.} $\loki$ improves baseline for all settings of $K$. 
We summarize the metric space graph by generating 10,000 supernodes. 
} \label{table:lshtc}
\begin{center}
    \begin{tabular}{r|cc}
    \toprule 
    $K$ & 5-NN  & 5-NN + \loki   \\
    \midrule
    \rowcolor[gray]{.9} 
    50  & 1.2519   &{\bf 0.2896} \\
    100 & 1.4476   &{\bf 0.3467} \\
    \rowcolor[gray]{.9} 
    150 & 1.7225  &{\bf 0.3969} \\
    200 & 1.7092  &{\bf 0.3171} \\
    \rowcolor[gray]{.9} 
    250 & 1.6465  &{\bf 0.3995} \\
    300 & 1.6465  &{\bf 0.4004} \\
    \bottomrule
    \end{tabular}
\end{center}
\end{wraptable}

\subsection{$\loki$ on Partially-Observed Label Spaces}

To validate our approach on partially observed label spaces, we evaluate the performance of adapting a logistic classifier trained on SimCLR embeddings of ImageNet \cite{imagenet}, 5-NN models trained on a 9,419 class subset of the PubMed dataset,\footnote{\url{https://www.kaggle.com/datasets/owaiskhan9654/pubmed-multilabel-text-classification}} and the 325,056-class LSHTC dataset \cite{Partalas2015LSHTCAB}. 

\paragraph{Setup} 
For ImageNet, we use the WordNet phylogenetic tree as the metric space \cite{barz2019hierarchy}. 
In this setting, we sample random subsets of size $K$ of the 1000 ImageNet classes and compare a baseline one-vs-rest classifier to the same classifier but using $\loki$ to adapt predictions to classes beyond the original $K$. 
We conduct two sets of experiments. 
In the first, we sample $K$ classes uniformly, while in the second, we adopt a more realistic sampler---we sample from a Gibbs distribution: $P(\lambda | \theta) = \frac{1}{Z}\exp(-\theta d(\lambda, \lambda^c))$,
where $\lambda^c = m_{\set{Y}}(\boldsymbol{1}_N)$ is the centroid of the metric space, $\theta$ is the concentration parameter around the centroid, and $Z$ is the normalizer. 
While the Gibbs distribution sampler is more realistic, it is also the more challenging setting---classes which have low probability according to this distribution are less likely to appear in the locus. 
For PubMed, we derive our metric from Euclidean distances between SimCSE class embeddings \cite{gao2021simcse}. 
Finally for LSHTC, we summarize the default graph by randomly selecting nodes and merging them with their neighbors until we obtain a graph with 10,000 supernodes representing sets of classes. 

\begin{figure}[t!]
    \centerline{
        \includegraphics[width=.4\columnwidth]{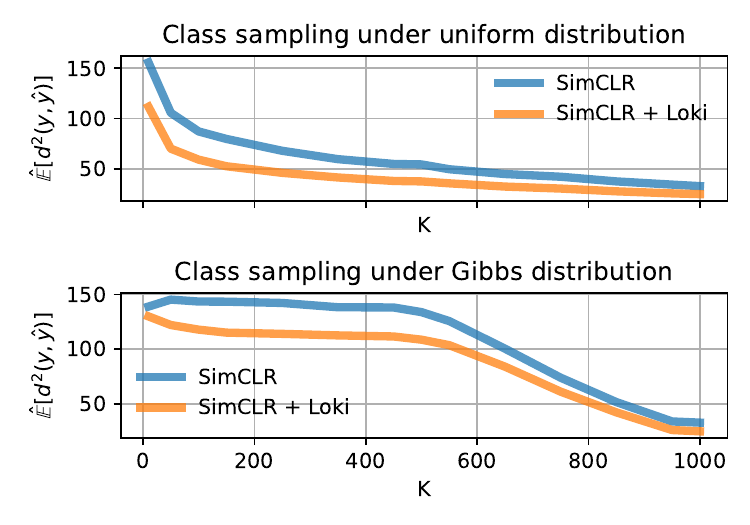}
        \includegraphics[width=.4\columnwidth]{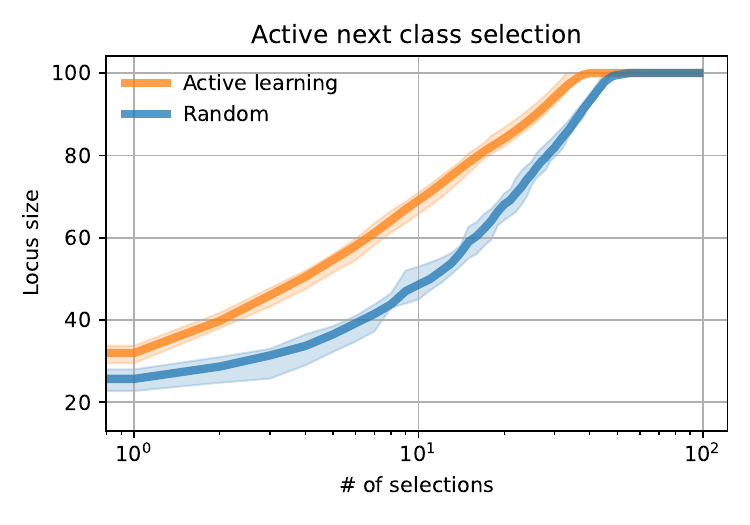} 
    }
    \centerline{
        \includegraphics[width=.8\columnwidth]{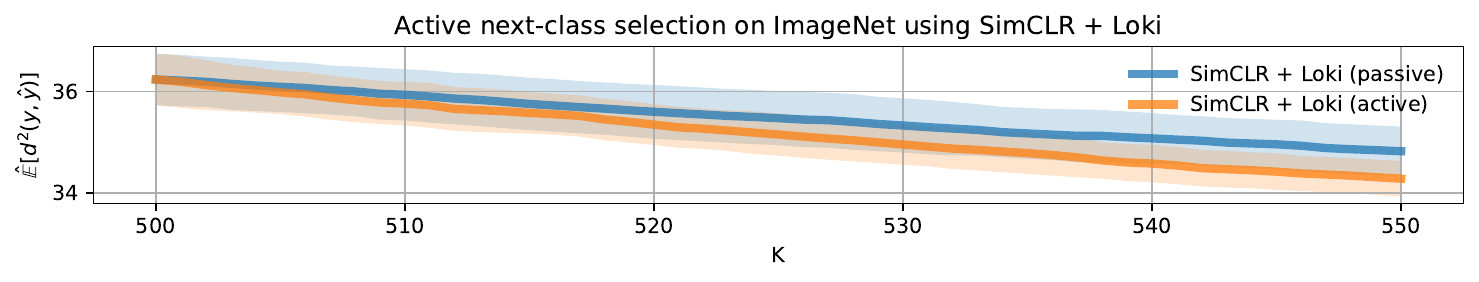}
    }
    \caption{(Top left) \textbf{ImageNet.} Mean squared distances under uniform class sampling and under the Gibbs distribution---$\loki$ improves upon the baseline SimCLR one-vs-rest classifier. 
    (Top right) \textbf{Synthetic.} Active class selection consistently leads to larger loci compared to uniform sampling.
    (Bottom) \textbf{Active selection on ImageNet.} Active class selection improves performance on ImageNet. 
    }
    \label{fig:imagenet_rel_improve}
\end{figure}

\paragraph{Results} Figure~\ref{fig:imagenet_rel_improve} shows the mean squared distances compared to the baseline one-vs-rest classifiers, across various settings of $K$. 
We find that $\loki$ always significantly outperforms the baseline, even in the more challenging setting of sampling according to a Gibbs distribution. 
Tables~\ref{table:pubmed}~and~\ref{table:lshtc} show our improvements when using $\loki$ over the baseline 5-NN models. 
While $\loki$ consistently yields an improvement on PubMed and LSHTC, the improvement is more dramatic on LSHTC. 


\subsection{Large Loci via Active Learning}
We evaluate our active next class selection approach on increasing the locus size. 
We expect that compared to passive (random) selection, our active approach will lead to larger loci, and in practice, a larger set of possible classes that can be predicted using $\loki$ while observing fewer classes during training. 

\paragraph{Setup} We compare with a baseline that randomly selects the next class. 
We first generate a synthetic random tree with size $N$ and fix an initial $K$. In active selection, we use the approach described in Theorem~\ref{thm:active_largest} to select the next node. 
As a passive baseline, we randomly sample (without replacement) the remaining nodes that are not in $\Lambda$.

\paragraph{Results} The results are shown in Figure~\ref{fig:imagenet_rel_improve}. 
We set $N = 100$ and the initial $K = 3$, then we average the result over 10 independent trials.
The active approach consistently outperforms the random baseline as it attains a larger locus with fewer nodes selected.


\subsection{Improving Performance via Active Learning}
Next, we evaluate the our active next class selection approach on improving error on ImageNet. 
While we found that the the active approach indeed leads to an increased locus size compared to the passive baseline, we expect that this increased locus size will lead to improved performance. 

\paragraph{Setup}
We randomly sample 500 ImageNet classes, and using the WordNet metric space, we use our active approach to iteratively sample 50 more classes. 
We compare this to a passive baseline in which the 50 classes are sampled randomly. 
We repeat this experiment over 10 independent trials.
{Theorem~\ref{thm:active_largest} applies to trees in which every vertex is a label. In this leaf-only WordNet experiment, we evaluate the selection strategy empirically without an optimality guarantee.} 

\paragraph{Results}
Figure~\ref{fig:imagenet_rel_improve} shows that our active approach yields improved error over the passive baseline. 
The gap between the active approach and the passive baseline widens with more selection rounds.

\begin{figure}[t!]
    \centerline{
        \includegraphics[width=.6\columnwidth]{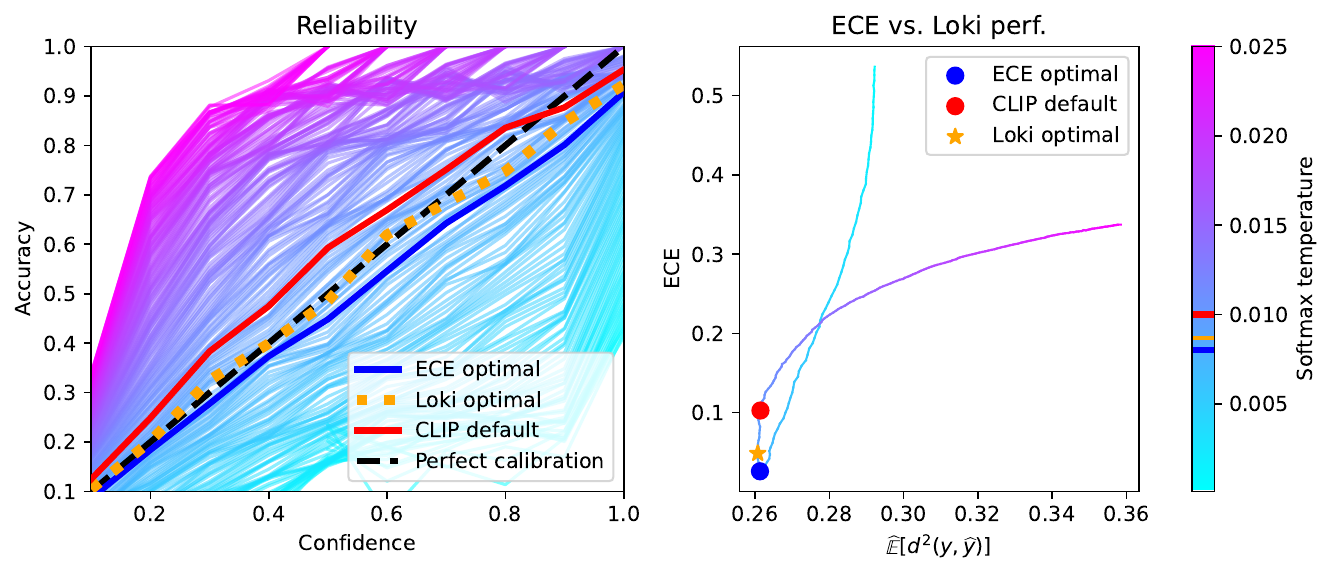}
        \includegraphics[width=.4\columnwidth]{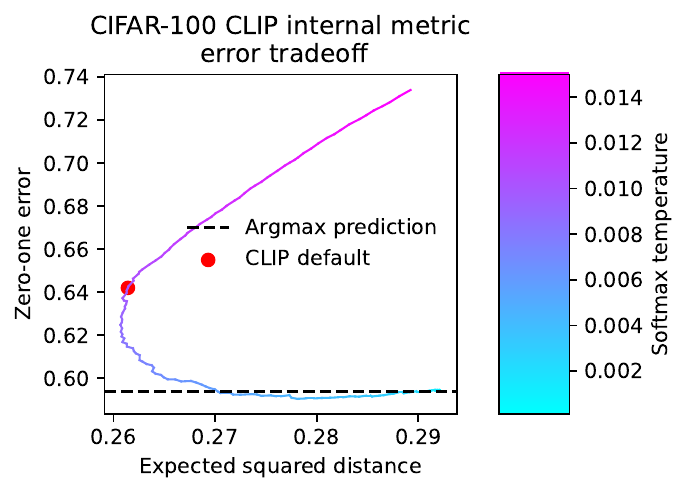} 
    }
    \caption{
    \textbf{CLIP on CIFAR-100 with no external metric.} 
    (Left) Reliability diagrams across a range of Softmax temperatures, highlighting the CLIP default temperature, the optimal temperature for $\loki$, and the minimizer of the Expected Calibration Error (ECE). All three are well-calibrated. 
    (Center) Tradeoff between optimizing for ECE and the expected squared distance. As with the reliability diagrams, the CLIP default temperature, the $\loki$-optimal temperature, and the ECE-optimal temperature are similar. 
    (Right) Tradeoff between optimizing for zero-one error and the expected squared distance. Temperature can be tuned to improve \textit{accuracy} when using $\loki$. 
    }
    \label{fig:calibration}
\end{figure}

\subsection{Improving Accuracy via Calibration}
Finally, we evaluate the effect of calibrating the Softmax outputs on the performance of $\loki$. 

\paragraph{Setup}
We calibrate via Softmax temperature scaling \cite{pmlr-v70-guo17a} using CLIP on CIFAR-100. 
We do not use an external metric space, and instead use Euclidean distance applied to the CLIP text encoder. 

\paragraph{Results}
The reliability diagram in Figure~\ref{fig:calibration} shows that the optimal Softmax temperature for $\loki$ is both close to the default temperature used by CLIP and to the optimally-calibrated temperature. 
In Figure~\ref{fig:calibration} (right), we find that appropriate tuning of the temperature parameter \textit{can lead to improved accuracy with CLIP}, even when no external metric space is available. 


\section{Conclusion}
In this work, we proposed $\loki$---a simple adaptor for pretrained models to enable the prediction of additional classes that are unobserved during training by using metric space information. 
{We studied finite-sample estimation error and excess prediction risk, optimal label subsets for several metric structures, and an active selection strategy.} 
Experimentally, we showed that $\loki$ can be used to improve CLIP even without external metric space information, can be used to predict a large number of unobserved classes, and a validation of our active selection strategy. 

\section*{Acknowledgements}
We are grateful for the support of the NSF under CCF2106707 (Program Synthesis for Weak Supervision) and the Wisconsin Alumni Research Foundation (WARF).

%
%
%
%
%
\bibliography{main}
\bibliographystyle{plainnat}
\newpage
\appendix
\onecolumn

The appendix is organized as follows: in Appendix~\ref{app:deferred_proofs}, we provide proofs of the claims made in the main text. 
Next, in Appendix~\ref{app:time_complexity}, we provide algorithms and time complexity analyses for the algorithms referenced in the main text. 
Appendix~\ref{app:experiments} contains additional experimental results. 
In Appendix~\ref{app:experimental_details}, we provide additional details about our experimental setup. 
Then in Appendix~\ref{app:broader_impacts}, we discuss the broader impacts and limitations of $\loki$. 
Finally, in Appendix~\ref{app:vis}, we provide more details and examples of the locus visualizations shown in Figure~\ref{fig:banner}.

\section{Deferred Proofs}
\label{app:deferred_proofs}

\begingroup
\subsection{Proof of Theorem~\ref{theorem:learning} ($\loki$ sample complexity)}
We first give bounds on probability estimation and its effect on the Fr\'echet objective and prediction risk. We then give bounds under a margin assumption. Theorem~\ref{thm:gaussian_covering} and the estimation and covering arguments that follow prove Theorem~\ref{theorem:learning}.

\begin{definition}[Fr\'echet objective]
For a finite metric space $(\mathcal Y,d)$, a nonempty subset $\Lambda=\{y_1,\ldots,y_K\}$, and a probability vector $p$ on $\Lambda$, define
\[
\Psi_p(z)=\sum_{i=1}^K p_i d^2(z,y_i),
\qquad y_p\in m_\Lambda(p)=\arg\min_{z\in\mathcal Y}\Psi_p(z).
\]
Use a fixed ordering of $\mathcal Y$ to break ties throughout.
\end{definition}

\begin{lemma}[Stability of the Fr\'echet objective]
\label{lemma:frechet_stability}
If $D=\operatorname{diam}_d(\mathcal Y)$ and $p,q$ are probability vectors on the same support $\Lambda\subseteq\mathcal Y$, then
\[
0\le \Psi_p(y_q)-\Psi_p(y_p)\le D^2\|p-q\|_1.
\]
\end{lemma}
\begin{proof}
Set $a_i=d^2(y_q,y_i)-d^2(y_p,y_i)$. Both squared distances lie in $[0,D^2]$, so $|a_i|\le D^2$. Optimality of $y_q$ gives $\sum_iq_ia_i\le0$. Thus
\[
0\le\sum_i p_i a_i
\le\sum_i(p_i-q_i)a_i
\le D^2\sum_i|p_i-q_i|.
\]
The first inequality follows from optimality of $y_p$.
\end{proof}

\begin{theorem}[$\loki$ finite-sample objective and excess-risk bounds]
\label{thm:learning_formal}
Let $(\mathcal Y,d)$ be a finite metric space, $N=|\mathcal Y|\ge1$, $D=\operatorname{diam}_d(\mathcal Y)$, and $\emptyset\ne\Lambda=\{y_1,\ldots,y_K\}\subseteq\mathcal Y$. Let $d\ge1$, $R\ge0$, and assign each label $y\in\mathcal Y$ a feature mean $\mu_y\in\mathbb R^d$ with $\|\mu_y\|_2\le R$. Write $\mu_i=\mu_{y_i}$ for observed labels. The scalar $d$ denotes feature dimension, whereas $d(\cdot,\cdot)$ denotes the label metric, as in the main text. The label metric need not agree with distances between feature means.
The $n$ training pairs $(x^{(j)},y^{(j)})$ are independent and identically distributed, with $y^{(j)}$ uniform on $\Lambda$ and $x^{(j)}\mid y^{(j)}=y\sim\mathcal N(\mu_y,I_d)$. Independently, a test pair $(x,y_*)$ has $y_*$ uniform on $\mathcal Y$ and $x\mid y_*=y\sim\mathcal N(\mu_y,I_d)$.
Write $n_i=|\{j:y^{(j)}=y_i\}|$ and set
\[
\widehat\mu_i=\frac1{n_i}\sum_{j:y^{(j)}=y_i}x^{(j)}\quad(n_i>0),
\qquad \widehat\mu_i=0\quad(n_i=0).
\]
Write $\mathbb P_i(x)=\mathbb P(y_i\mid x)$ and $\widehat{\mathbb P}_i(x)=\hat{\mathbb P}(y_i\mid x)$ for the exact and estimated observed-label probabilities, and let $\mathbf P(x)=[\mathbb P_i(x)]_{i\in[K]}$ and $\widehat{\mathbf P}(x)=[\widehat{\mathbb P}_i(x)]_{i\in[K]}$. The exact probabilities condition on label membership in $\Lambda$. With known identity covariance and equal priors,
\[
\mathbb P_i(x)=\frac{\exp(-\|x-\mu_i\|_2^2/2)}{\sum_{j=1}^K\exp(-\|x-\mu_j\|_2^2/2)},
\qquad
\widehat{\mathbb P}_i(x)=\frac{\exp(-\|x-\widehat\mu_i\|_2^2/2)}{\sum_{j=1}^K\exp(-\|x-\widehat\mu_j\|_2^2/2)}.
\]
Let $\Psi_p(z)=\sum_i p_i d^2(z,y_i)$ and $\hat y_*(x)=y_{\widehat{\mathbf P}(x)}$, with fixed tie breaking. Define the conditional test risk and Bayes risk by
\[
\mathcal R(\hat y_*)=\mathbb E[d^2(y_*,\hat y_*(x))\mid\mathcal D_{\rm train}],
\qquad
\mathcal R^*=\inf_{f:\mathbb R^d\to\mathcal Y\ {\rm measurable}}
\mathbb E\,d^2(y_*,f(x)).
\]
For $0<\delta<1$ and $n\ge8K\log(2K/\delta)$, put
\[
\epsilon_n=\sqrt{\frac{4dK\log(4dK/\delta)}{n}},
\qquad
\eta_n=\min\{2,\;2\epsilon_n(2R+\sqrt d)+\epsilon_n^2\}.
\]
With probability at least $1-\delta$ over the training data, all of the following hold:
\begin{align}
\min_i n_i&\ge\frac{n}{2K},\qquad
\max_i\|\widehat\mu_i-\mu_i\|_2\le\epsilon_n;
\label{eq:gaussian_mean_bound}\\
\mathbb E_x\|\widehat{\mathbf P}(x)-\mathbf P(x)\|_1&\le\eta_n;
\label{eq:gaussian_posterior_bound}\\
\mathbb E_x\left[\Psi_{\mathbf P(x)}(\hat y_*(x))-\min_{z\in\mathcal Y}\Psi_{\mathbf P(x)}(z)\right]
&\le D^2\eta_n;
\label{eq:gaussian_objective_bound}\\
\mathcal R(\hat y_*)&\le\mathcal R^*+2D^2\left(1-\frac KN\right)+D^2\eta_n.
\label{eq:gaussian_risk_bound}
\end{align}
Here $\mathbb E_x$ uses the test marginal, conditional on the training data.
\end{theorem}

\begin{proof}
\textbf{Class counts and mean estimation.}
Each $n_i$ is binomial with mean $n/K$. The multiplicative Chernoff bound and a union bound give
\[
\Pr\!\left(\min_i n_i<\frac{n}{2K}\right)
\le K\exp\!\left(-\frac{n}{8K}\right)\le\frac\delta2.
\]
Conditional on the training labels, each coordinate of $\widehat\mu_i-\mu_i$ is Gaussian with variance $1/n_i$. On the count event, the Gaussian tail bound and a union bound over the $dK$ coordinates imply
\[
\Pr\!\left(\max_{i,\ell}|(\widehat\mu_i-\mu_i)_\ell|>
\frac{\epsilon_n}{\sqrt d}\ \middle|\ y^{(1)},\ldots,y^{(n)}\right)
\le 2dK\exp\!\left(-\frac{n\epsilon_n^2}{4dK}\right)=\frac\delta2.
\]
The coordinate bound implies the Euclidean feature-mean bound in~\eqref{eq:gaussian_mean_bound}. Combining the two failure probabilities gives an event of probability at least $1-\delta$. Fix any training sample in this event for the remainder of the proof.

\textbf{Posterior perturbation.}
The softmax map $\sigma_i(s)=e^{s_i}/\sum_j e^{s_j}$ satisfies
\[
\|\sigma(s)-\sigma(t)\|_1\le2\|s-t\|_\infty.
\]
Indeed, along the segment $u(a)=t+a(s-t)$, writing $v=s-t$, its derivative is
\[
\frac{d}{da}\sigma_i(u(a))
=\sigma_i(u(a))\left(v_i-\sum_j\sigma_j(u(a))v_j\right).
\]
Summing absolute values bounds the derivative's $\ell_1$ norm by $2\|v\|_\infty$; integration over $a\in[0,1]$ proves the claim.
For $s_i(x)=-\|x-\mu_i\|_2^2/2$ and $\widehat s_i(x)=-\|x-\widehat\mu_i\|_2^2/2$, expanding the squares gives, for every $x\in\mathbb R^d$,
\begin{align*}
|\widehat s_i(x)-s_i(x)|
&\le \|x-\mu_i\|_2\|\widehat\mu_i-\mu_i\|_2
       +\tfrac12\|\widehat\mu_i-\mu_i\|_2^2\\
&\le\epsilon_n(\|x\|_2+R)+\tfrac12\epsilon_n^2.
\end{align*}
Consequently,
\[
\|\widehat{\mathbf P}(x)-\mathbf P(x)\|_1
\le2\epsilon_n(\|x\|_2+R)+\epsilon_n^2.
\]
Under the test model, $x=\mu_{y_*}+Z$ with $Z\sim\mathcal N(0,I_d)$ independent of $y_*$, so
\[
\mathbb E\|x\|_2\le R+\mathbb E\|Z\|_2
\le R+\sqrt{\mathbb E\|Z\|_2^2}=R+\sqrt d.
\]
Taking expectations and using the universal bound $\|\widehat{\mathbf P}-\mathbf P\|_1\le2$ proves~\eqref{eq:gaussian_posterior_bound}. Lemma~\ref{lemma:frechet_stability}, applied at each $x$ and then averaged, gives~\eqref{eq:gaussian_objective_bound}.

\textbf{Full-test excess risk.}
Let $r_y(x)=\Pr(y_*=y\mid x)$ be the full test posterior on $\mathcal Y$, and let $\mathbf P^{\rm ext},\widehat{\mathbf P}^{\rm ext}$ extend $\mathbf P,\widehat{\mathbf P}$ by zero outside $\Lambda$. Because $\mathcal Y$ is finite, a fixed tie-breaking minimizer of $\Psi_{r(x)}$ is a measurable Bayes rule. Applying Lemma~\ref{lemma:frechet_stability} on support $\mathcal Y$ therefore gives
\[
\mathcal R(\hat y_*)-\mathcal R^*
\le D^2\mathbb E_x\|r(x)-\widehat{\mathbf P}^{\rm ext}(x)\|_1.
\]
Write $t(x)=\sum_{y\in\Lambda}r_y(x)>0$. Equal priors and the common Gaussian likelihood imply $\mathbb P_i(x)=r_{y_i}(x)/t(x)$. Thus
\begin{align*}
\|r(x)-\mathbf P^{\rm ext}(x)\|_1
&=\sum_{i=1}^K\left(\frac{r_{y_i}(x)}{t(x)}-r_{y_i}(x)\right)
+\sum_{y\notin\Lambda}r_y(x)\\
&=2(1-t(x)).
\end{align*}
By conditional expectation, $\mathbb E_x t(x)=\Pr(y_*\in\Lambda)=K/N$. The triangle inequality and~\eqref{eq:gaussian_posterior_bound} yield
\[
\mathbb E_x\|r(x)-\widehat{\mathbf P}^{\rm ext}(x)\|_1
\le2\left(1-\frac KN\right)+\eta_n,
\]
which proves~\eqref{eq:gaussian_risk_bound}.
\end{proof}

\paragraph{Interpretation and scope.}
Both the deterministic lemma and the finite-sample theorem allow any finite label metric $d$, including shortest-path distance on a finite connected graph. The Gaussian assumption concerns only the feature space $\mathbb R^d$ and the stated supervised estimator. The label diameter $D$ and feature-mean radius $R$ are separate parameters; no relation such as $D\le2R$ is assumed. It requires neither a net nor a positive lower sensitivity constant. Its estimation term is $O(n^{-1/2})$ for fixed $d,K,R,\delta$. The support term is an upper bound on the discrepancy caused by conditioning on observed labels, and can be loose; it does not guarantee accurate prediction of every unseen class. When $K=N$, the bound gives convergence to Bayes risk. That risk need not vanish: for $\mathcal Y=\Lambda=\{-1,+1\}$ with $d(u,v)=|u-v|$, feature means $\mu_y=y$, and unit Gaussian noise, the Bayes rule is the sign of $x$ and its squared-distance risk is $4\Phi(-1)>0$, where $\Phi$ is the standard normal distribution function. Convergence to an exact-probability Fr\'echet decision must therefore be distinguished from recovery of the random test label.

\begin{corollary}[Prediction stability under a margin condition]
\label{cor:margin_prediction}
Under the setup and sample-size requirement of Theorem~\ref{thm:learning_formal}, suppose $N\ge2$. Let $y_*^{\circ}(x)=y_{\mathbf P(x)}$ with the same fixed tie-breaking rule, and define
\[
g(x)=\min_{z\in\mathcal Y\setminus\{y_*^{\circ}(x)\}}
\bigl[\Psi_{\mathbf P(x)}(z)-\Psi_{\mathbf P(x)}(y_*^{\circ}(x))\bigr],
\qquad A(x)=1+\|x\|_2+R.
\]
Assume that a constant $C>0$, independent of the training sample and $n$, satisfies the following condition under the test marginal:
\begin{equation}
\Pr_x\bigl(g(x)\le tA(x)\bigr)\le Ct
\quad\text{for all }t>0.
\label{eq:margin_condition}
\end{equation}
Set $b_n=\min\{1,\;CD^2(2\epsilon_n+\epsilon_n^2)\}$. With probability at least $1-\delta$ over training, simultaneously with the conclusions of Theorem~\ref{thm:learning_formal},
\begin{align}
\Pr_x\bigl(\hat y_*(x)\ne y_*^{\circ}(x)\bigr)&\le b_n,
\label{eq:margin_disagreement}\\
\mathbb E_x\,d^2(\hat y_*(x),y_*^{\circ}(x))&\le D^2b_n,
\label{eq:margin_distance}\\
\bigl|\mathcal R(\hat y_*)-\mathcal R(y_*^{\circ})\bigr|&\le D^2b_n.
\label{eq:margin_risk}
\end{align}
All probabilities and expectations involving the fitted predictor are conditional on training, and $\mathcal R(y_*^{\circ})=\mathbb E\,d^2(y_*,y_*^{\circ}(x))$. In particular, writing $\hat y_*$ for the prediction $\hat y_*(x)$ at the test input,
\[
\mathbb E\,d^2(y_*,\hat y_*)\le\mathbb E\,d^2(y_*,y_*^{\circ}(x))+CD^4(2\epsilon_n+\epsilon_n^2).
\]
\end{corollary}

\begin{proof}
The gap $g$ is nonnegative. Condition~\eqref{eq:margin_condition} implies $\Pr(g(x)=0)=0$, since $\{g(x)=0\}\subseteq\{g(x)\le tA(x)\}$ for every $t>0$. Thus the exact minimizer is unique almost surely, although the fixed tie-breaking convention defines it everywhere.

Fix a training sample in the event of Theorem~\ref{thm:learning_formal}. Its posterior-perturbation argument gives, for every $x$,
\[
\|\widehat{\mathbf P}(x)-\mathbf P(x)\|_1
\le2\epsilon_n(\|x\|_2+R)+\epsilon_n^2
\le(2\epsilon_n+\epsilon_n^2)A(x),
\]
where the last inequality uses $A(x)\ge1$ and $A(x)\ge\|x\|_2+R$. By Lemma~\ref{lemma:frechet_stability},
\[
\Psi_{\mathbf P(x)}(\hat y_*(x))-\Psi_{\mathbf P(x)}(y_*^{\circ}(x))
\le D^2(2\epsilon_n+\epsilon_n^2)A(x).
\]
Whenever $\hat y_*(x)\ne y_*^{\circ}(x)$, the left-hand side is at least $g(x)$ by its definition. Therefore
\[
\{\hat y_*(x)\ne y_*^{\circ}(x)\}
\subseteq
\{g(x)\le D^2(2\epsilon_n+\epsilon_n^2)A(x)\}.
\]
Here $D>0$ because $N\ge2$ and the labels are distinct. Applying~\eqref{eq:margin_condition} at $t=D^2(2\epsilon_n+\epsilon_n^2)>0$, and capping the probability at one, proves~\eqref{eq:margin_disagreement}.

Both predictions lie in the metric space $(\mathcal Y,d)$ of diameter $D$, so
\[
d^2(\hat y_*(x),y_*^{\circ}(x))
\le D^2\mathbf 1\{\hat y_*(x)\ne y_*^{\circ}(x)\}.
\]
Taking expectations proves~\eqref{eq:margin_distance}. Likewise, for every test pair,
\[
\left|d^2(y_*,\hat y_*(x))-d^2(y_*,y_*^{\circ}(x))\right|
\le D^2\mathbf 1\{\hat y_*(x)\ne y_*^{\circ}(x)\},
\]
because both squared losses belong to $[0,D^2]$ and coincide when the predictions agree. Taking expectations and then bounding the absolute difference of risks proves~\eqref{eq:margin_risk}. All steps use the same training event, so no additional failure probability is incurred.
\end{proof}

\paragraph{Role of the margin assumption.}
Condition~\eqref{eq:margin_condition} controls how frequently the exact objective nearly ties, allowing arbitrarily small positive gaps. The factor $A(x)$ matches the input dependence of the posterior perturbation and avoids any bounded-input assumption for Gaussian observations. For fixed $C,d,K,R,D,\delta$, the prediction-distance and risk-difference bounds are $O(n^{-1/2})$. When $\Lambda=\mathcal Y$, $y_*^{\circ}$ is a Bayes rule and $\mathcal R(y_*^{\circ})=\mathcal R^*$. For a proper subset, its risk reflects the behavior of exact-probability $\loki$ on both observed and unseen labels; bounding that baseline requires additional information about the geometry and noise. Neither a locus cover nor the Gaussian model alone is asserted to imply the margin condition. If $N=1$, both predictors equal the sole label and the distance and risks are identically zero, without a margin assumption.

\paragraph{Alternative margin-based prediction bound.}
To expose the feature-radius factor, the margin-based bound uses the normalized input scale
\[
B_R(x)=1+\frac{\|x\|_2}{1+R},\qquad
A(x)=(1+R)B_R(x)=1+\|x\|_2+R.
\]
Its margin assumption is
\begin{equation}
\Pr_x\bigl(g(x)\le tD^2 B_R(x)\bigr)\le t/\alpha
\quad(t>0),\qquad 0<\alpha\le1.
\label{eq:normalized_margin}
\end{equation}
Substituting $t=u(1+R)/D^2$ gives condition~\eqref{eq:margin_condition} with $C=(1+R)/(\alpha D^2)$. Here $D>0$ because $N\ge2$. This normalization makes the factor $1+R$ explicit rather than including it in the margin parameter.

We prove the following explicit form of the margin-based bound on the same training event, writing $\mathcal E_\Lambda=\mathbb E\,d^2(y_*,y_*^{\circ}(x))$:
\begin{equation}
\left|\mathbb E\,d^2(y_*,\hat y_*)-\mathcal E_\Lambda\right|
\le \min\!\left\{D^2,\;
\frac{8D^2(1+R)}{\alpha}\sqrt{\frac{dK\log(4dK/\delta)}{n}}\right\}.
\label{eq:normalized_margin_risk}
\end{equation}
All expectations involving $\hat y_*$, the prediction at test input $x$, are conditional on training. This bounds the squared distance to the actual test label $y_*$.

\begin{proof}
Let $a_n=dK\log(4dK/\delta)/n$, so that $\epsilon_n=2\sqrt{a_n}$. On the training event of probability at least $1-\delta$ in Theorem~\ref{thm:learning_formal}, apply Corollary~\ref{cor:margin_prediction} with $C=(1+R)/(\alpha D^2)$. Equation~\eqref{eq:margin_risk} gives
\begin{align*}
\left|\mathbb E\,d^2(y_*,\hat y_*)-\mathcal E_\Lambda\right|
&\le D^2\min\left\{1,\frac{(1+R)(2\epsilon_n+\epsilon_n^2)}{\alpha}\right\}\\
&=D^2\min\left\{1,\frac{(1+R)(4\sqrt{a_n}+4a_n)}{\alpha}\right\}.
\end{align*}
If $a_n\le1$, then $a_n\le\sqrt{a_n}$, so this is at most $8D^2(1+R)\sqrt{a_n}/\alpha$. If $a_n>1$, it is at most $D^2\le D^2(1+R)\sqrt{a_n}/\alpha$, since $R\ge0$ and $\alpha\le1$. It is also at most $D^2$ in both cases. This proves~\eqref{eq:normalized_margin_risk}; taking the upper direction yields a margin-based prediction-error bound with universal big-$O$ constant $8$. The same upper bound applies to $\mathbb E_x\,d^2(\hat y_*(x),y_*^{\circ}(x))$ by~\eqref{eq:margin_distance}.
\end{proof}

\paragraph{A fixed configuration with varying radius.}
Fix the metric label space, the observed subset, and vectors $\{v_y\}_{y\in\mathcal Y}\subseteq B_2(0,1)$. For each $R>0$, consider the model with $\mu_y(R)=Rv_y$ and unit Gaussian noise. Write $\mathbb E_R$ for its test expectation and $\alpha_R$ for any margin parameter satisfying~\eqref{eq:normalized_margin} in that model. The preceding proof applies at each fixed $R$ and gives
\[
\mathbb E_R[d^2(y_*,\hat y_*)]
\le\mathcal E_\Lambda(R)+
\frac{8D^2(1+R)}{\alpha_R}\sqrt{\frac{dK\log(4dK/\delta)}{n}}
\]
with training probability at least $1-\delta$. If $\alpha_R\ge\underline\alpha>0$ for every radius in a specified range, replacing $\alpha_R$ by $\underline\alpha$ gives a common explicit upper bound on the estimation term. This is a guarantee at each fixed radius, not a simultaneous guarantee for training samples at all radii.

The scaling fixes the feature configuration up to dilation: $x/R\mid y\sim\mathcal N(v_y,R^{-2}I_d)$. Thus varying $R$ changes feature separation relative to noise, while keeping the label loss fixed. The margin parameters and oracle errors can change with that ratio. No uniform positive margin bound or monotonicity of $\mathcal E_\Lambda(R)$ is implied merely by the radius constraint; these need separate geometric justification.

\paragraph{Interpretation.}
Under the margin assumption, the bound gives an $O(n^{-1/2})$ difference between the fitted predictor's squared-distance risk and $\mathcal E_\Lambda$. The parameter $\alpha$ is defined by~\eqref{eq:normalized_margin}. The baseline $\mathcal E_\Lambda$ accounts for noise and the observed label set and is independent of training size. For $N=1$, the loss is zero without a margin assumption.

\subsubsection{Gaussian estimation and covering-radius bounds}
We bound the prediction error using estimates of the class means, localization of the Gaussian weights, and a covering of the feature means. The baseline accounts for noise and the observed label set.

\begin{theorem}[Gaussian localization and true-label loss]
\label{thm:gaussian_covering}
Use the metric label space, Gaussian feature model, and independent training/test distributions of Theorem~\ref{thm:learning_formal}. Write $\mu_i=\mu_{y_i}$. Allow any estimated means $\widehat\mu_i$ and their equal-prior Gaussian posterior $\widehat{\mathbf P}(x)$, and suppose
\[
\max_{i\in[K]}\|\widehat\mu_i-\mu_i\|_2\le\varepsilon.
\]
Define
\[
h=\max_{y\in\mathcal Y}\min_{i\in[K]}\|\mu_y-\mu_i\|_2,
\quad H=\sqrt d+h+\sqrt{2\log K},
\quad \eta_h=\min\{2,\;2\varepsilon H+\varepsilon^2\},
\]
and let $\mathcal C_\Lambda=4\mathbb E\Psi_{\mathbf P(x)}(y_*)$. Conditional on these estimated means,
\begin{align}
\mathbb E_x\|\widehat{\mathbf P}(x)-\mathbf P(x)\|_1
&\le\eta_h,\label{eq:net_posterior}\\
\mathbb E[d^2(y_*,\hat y_*)]
&\le\min\{D^2,\;\mathcal C_\Lambda+2D^2\eta_h\}.
\label{eq:net_true_loss}
\end{align}
If there is an $L>0$ such that $d(y,z)\le L\|\mu_y-\mu_z\|_2$ for all $y,z\in\mathcal Y$, set $G=2\sqrt d+h+\sqrt{2\log K}$. Then
\begin{equation}
\mathcal C_\Lambda\le4L^2G^2,
\qquad
\mathbb E[d^2(y_*,\hat y_*)]\le\min\{D^2,\;4L^2(G+2\varepsilon)^2\}.
\label{eq:net_explicit_baseline}
\end{equation}
The expectations are over an independent test pair; the conclusions hold for each fixed set of estimated means satisfying the error bound.
\end{theorem}

\begin{proof}
\textbf{Gaussian weights and distance to the nearest mean.}
For any vectors $\nu_1,\ldots,\nu_K$, let $r_i=\|x-\nu_i\|_2$, $\ell=\min_i r_i^2$, and $q_i=e^{-r_i^2/2}/\sum_j e^{-r_j^2/2}$. Put $Z_0=\sum_i e^{-(r_i^2-\ell)/2}\ge1$. The entropy identity gives
\[
-\sum_iq_i\log q_i
=\frac12\sum_iq_i(r_i^2-\ell)+\log Z_0.
\]
By concavity of the logarithm, $-\sum_iq_i\log q_i\le\log K$. Consequently,
\begin{equation}
\sum_iq_i r_i^2\le\min_i r_i^2+2\log K.
\label{eq:gaussian_localization}
\end{equation}
This controls the whole weighted sum, without converting a finite local count into a volume fraction.

\textbf{Perturbing the Gaussian means.}
Set $\Delta_i=\widehat\mu_i-\mu_i$, $\mu_i(t)=\mu_i+t\Delta_i$ for $0\le t\le1$, and let $p_i(t,x)$ be the posterior obtained from these interpolated means. Differentiating the normalized Gaussian weights yields
\[
\frac{\partial p_i(t,x)}{\partial t}
=p_i(t,x)\left(v_i(t,x)-\sum_jp_j(t,x)v_j(t,x)\right),
\qquad v_i(t,x)=\langle x-\mu_i(t),\Delta_i\rangle.
\]
Since $\|\Delta_i\|_2\le\varepsilon$, the triangle and Cauchy--Schwarz inequalities give
\[
\sum_i\left|\frac{\partial p_i(t,x)}{\partial t}\right|
\le2\varepsilon\sum_i p_i(t,x)\|x-\mu_i(t)\|_2
\le2\varepsilon\sqrt{\sum_i p_i(t,x)\|x-\mu_i(t)\|_2^2}.
\]
For a test pair write $x=\mu_{y_*}+Z$, where $Z\sim\mathcal N(0,I_d)$. Choose an observed mean within distance $h$ of $\mu_{y_*}$. Its interpolated version is within distance $h+t\varepsilon$, so
\[
\min_i\|x-\mu_i(t)\|_2\le\|Z\|_2+h+t\varepsilon.
\]
Applying~\eqref{eq:gaussian_localization} at time $t$ bounds the derivative norm by
\[
2\varepsilon\sqrt{(\|Z\|_2+h+t\varepsilon)^2+2\log K}
\le2\varepsilon(\|Z\|_2+h+t\varepsilon+\sqrt{2\log K}).
\]
Integrating from $0$ to $1$ proves the exact inequality
\[
\|\widehat{\mathbf P}(x)-\mathbf P(x)\|_1
\le2\varepsilon(\|Z\|_2+h+\sqrt{2\log K})+\varepsilon^2.
\]
Taking test expectations, using $\mathbb E\|Z\|_2\le\sqrt d$, and capping probability-vector distance at $2$ proves~\eqref{eq:net_posterior}. No Taylor remainder or matching of nearest-mean indices is needed.

\textbf{Comparing the fitted prediction with the actual label.}
For any probability vector $q$ and any $y\in\mathcal Y$, the metric triangle inequality gives
\[
d^2(y,y_q)\le2\sum_iq_i d^2(y,y_i)+2\sum_iq_i d^2(y_q,y_i)
=2\Psi_q(y)+2\Psi_q(y_q)\le4\Psi_q(y),
\]
where the last step uses optimality of $y_q$. Also, for any probability vectors $p,q$, their difference sums to zero, and $0\le d^2(y,y_i)\le D^2$. Therefore
\[
|\Psi_q(y)-\Psi_p(y)|
=\left|\sum_i(q_i-p_i)\left(d^2(y,y_i)-\frac{D^2}{2}\right)\right|
\le\frac{D^2}{2}\|q-p\|_1.
\]
Apply these inequalities with $y=y_*$, $q=\widehat{\mathbf P}(x)$, and $p=\mathbf P(x)$, then average over the test pair. Equation~\eqref{eq:net_posterior} gives~\eqref{eq:net_true_loss}; the cap follows from the diameter bound. The true label need not minimize either objective.

\textbf{An explicit baseline from feature geometry.}
Assume the stated compatibility bound with constant $L$. Weighted Minkowski, applied to
$\mu_{y_*}-\mu_i=(\mu_{y_*}-x)+(x-\widehat\mu_i)+(\widehat\mu_i-\mu_i)$, gives
\begin{align*}
\sqrt{\Psi_{\widehat{\mathbf P}(x)}(y_*)}
&\le L\left(\|Z\|_2+
\sqrt{\sum_i\widehat{\mathbb P}_i(x)\|x-\widehat\mu_i\|_2^2}+\varepsilon\right)\\
&\le L\left(\|Z\|_2+
\sqrt{(\|Z\|_2+h+\varepsilon)^2+2\log K}+\varepsilon\right)\\
&\le L(2\|Z\|_2+h+\sqrt{2\log K}+2\varepsilon).
\end{align*}
The second line uses~\eqref{eq:gaussian_localization} and the covering radius. Combining with $d^2(y_*,\hat y_*)\le4\Psi_{\widehat{\mathbf P}(x)}(y_*)$ and taking the $L^2$ norm over test noise yields
\[
\sqrt{\mathbb E[d^2(y_*,\hat y_*)]}
\le2L(2\sqrt d+h+\sqrt{2\log K}+2\varepsilon)=2L(G+2\varepsilon).
\]
This proves the risk bound in~\eqref{eq:net_explicit_baseline}. The same weighted-objective argument with exact means and $\varepsilon=0$ proves $\mathcal C_\Lambda\le4L^2G^2$.
\end{proof}

\paragraph{Median-of-means estimation and completion of Theorem~\ref{theorem:learning}.}
We estimate each class mean using a median-of-means estimator with the available examples from that class. Put $\ell=\log(4dK/\delta)$ and let $b$ be the smallest odd integer at least $4\ell$. For each class with $n_i\ge2b$, form $b$ disjoint blocks of $\lfloor n_i/b\rfloor$ examples, discard the remainder, compute their vector means, and take the coordinatewise median. Set the estimate to zero if $n_i<2b$. If $n\ge24K\ell$, then with probability at least $1-\delta$,
\begin{equation}
\max_i\|\widehat\mu_i-\mu_i\|_2
\le14\sqrt{\frac{dK\log(4dK/\delta)}{n}}.
\label{eq:net_mom}
\end{equation}
To prove this, the binomial count bound used in Theorem~\ref{thm:learning_formal} gives $n_i\ge n/(2K)$ for all $i$ except on probability at most $\delta/2$. Since $b<4\ell+2$ and $\ell\ge\log4$, the sample requirement implies $n\ge4Kb$, hence $n_i\ge2b$ on this event. Conditional on the labels, each block-coordinate mean has variance at most $2b/n_i$. Chebyshev's inequality bounds its probability of error exceeding $4\sqrt{b/n_i}$ by $1/8$. If a coordinate median exceeds this error, at least half of the $b$ independent blocks must fail. A union bound over subsets of blocks bounds that probability by
\[
2^b(1/8)^{b/2}\le e^{-b/4}.
\]
A further union bound over $dK$ coordinates gives failure probability at most $dKe^{-b/4}\le\delta/4$. On the resulting event every class-mean error is at most $4\sqrt{db/n_i}\le4\sqrt{2dKb/n}\le14\sqrt{dK\ell/n}$, proving~\eqref{eq:net_mom}. The per-class sample-mean estimator in Theorem~\ref{thm:learning_formal} also satisfies this bound under the same sample requirement.

Put $a_n=dK\log(4dK/\delta)/n$ and substitute $\varepsilon=14\sqrt{a_n}$ into Theorem~\ref{thm:gaussian_covering}. Its general metric conclusion is at most
\[
\mathcal C_\Lambda+56D^2H\sqrt{a_n}+392D^2a_n.
\]
Under feature compatibility the direct bound expands to
\[
4L^2(G+2\varepsilon)^2
=4L^2G^2+16L^2G\varepsilon+16L^2\varepsilon^2
\le4L^2G^2+224L^2G\sqrt{a_n}+3136L^2a_n.
\]
These establish intermediate bounds in terms of the actual covering radius, with universal constants and the stated training probability. The radius substitution below completes Theorem~\ref{theorem:learning}. For Euclidean label distances with $\mu_y=y$, take $L=1$. Other label metrics remain covered by~\eqref{eq:net_true_loss}; the compatibility bound is an additional assumption only for the explicit geometric baseline and sharper remainder.

\paragraph{Covering radius and sample complexity.}
Let the finite set of feature means lie in $B_2(0,R)$, with $R>0$, and let $h>0$. Choose a maximal subset whose distinct means are at distance greater than $h$, retaining one label for each selected mean. Write $M$ for its cardinality. Maximality ensures that every class mean lies within $h$ of a selected mean. Balls of radius $h/2$ about the selected means have disjoint interiors and lie inside a ball of radius $R+h/2$. Comparing their volumes gives
\[
M\left(\frac h2\right)^d\le\left(R+\frac h2\right)^d,
\qquad M\le\left(1+\frac{2R}{h}\right)^d.
\]
For any integer $2\le K\le|\mathcal Y|$, choose $h_K=2R/(K^{1/d}-1)$. The construction gives $M\le K$ selected labels and covering radius at most $h_K$. If $M<K$, add arbitrary remaining labels until there are exactly $K$. Adding labels cannot increase covering radius. Thus the same $K$ is both the actual observed-class count and the count used in the displayed covering guarantee; no lower bound on a count is substituted for an upper bound. If $R=0$, all feature means coincide, and any $K$ labels have covering radius zero, so the same conclusion holds.

For any observed subset satisfying this covering guarantee, the quantities in Theorem~\ref{thm:gaussian_covering} satisfy
\[
H\le\sqrt d+\frac{2R}{K^{1/d}-1}+\sqrt{2\log K},
\qquad
G\le2\sqrt d+\frac{2R}{K^{1/d}-1}+\sqrt{2\log K}.
\]
Every term in the intermediate bounds is nondecreasing in $H$ or $G$. Substitution therefore gives the following completely explicit general-metric bound:
\begin{align*}
\mathbb E[d^2(y_*,\hat y_*)]
&\le\mathcal C_\Lambda+
56D^2\left(\sqrt d+\frac{2R}{K^{1/d}-1}+\sqrt{2\log K}\right)
\sqrt{\frac{dK\log(4dK/\delta)}{n}}\\
&\quad+392D^2\frac{dK\log(4dK/\delta)}{n}.
\end{align*}
Under feature compatibility the corresponding explicit bound is
\begin{align*}
\mathbb E[d^2(y_*,\hat y_*)]
&\le4L^2\left(2\sqrt d+\frac{2R}{K^{1/d}-1}+\sqrt{2\log K}\right)^2\\
&\quad+224L^2\left(2\sqrt d+\frac{2R}{K^{1/d}-1}+\sqrt{2\log K}\right)
\sqrt{\frac{dK\log(4dK/\delta)}{n}}\\
&\quad+3136L^2\frac{dK\log(4dK/\delta)}{n}.
\end{align*}
\paragraph{Simplifying only the estimation term.}
We retain the sharper baseline above and absorb only the smaller $n^{-1}$ term into the $n^{-1/2}$ estimation term. For this calculation, write
\[
H_K=\sqrt d+\frac{2R}{K^{1/d}-1}+\sqrt{2\log K},\qquad
G_K=2\sqrt d+\frac{2R}{K^{1/d}-1}+\sqrt{2\log K},\qquad
a=\frac{dK\log(4dK/\delta)}{n}.
\]
The assumption $n\ge24K\log(4dK/\delta)$ implies $a\le d/24$. Since $H_K\ge\sqrt d$ and $G_K\ge2\sqrt d$,
\[
a\le\frac{H_K\sqrt a}{\sqrt{24}},\qquad
a\le\frac{G_K\sqrt a}{2\sqrt{24}}.
\]
The preceding general-metric bound is therefore at most
\[
\mathcal C_\Lambda+D^2\left(56+\frac{392}{\sqrt{24}}\right)H_K\sqrt a
\le\mathcal C_\Lambda+140D^2 H_K\sqrt a.
\]
Under feature compatibility it is at most
\[
4L^2G_K^2+L^2\left(224+\frac{3136}{2\sqrt{24}}\right)G_K\sqrt a
\le4L^2G_K^2+550L^2G_K\sqrt a.
\]
Substituting the displayed expressions for $H_K,G_K,$ and $a$ proves both compact displays in Theorem~\ref{theorem:learning}, with universal big-$O$ constants at most $140$ and $550$, respectively. The baseline $4L^2G_K^2$ is preserved exactly; only the vanishing estimation term is upper-bounded for brevity. All dependencies on $R,d,K,n,$ and $\delta$ remain explicit in the main statement, and the training probability remains at least $1-\delta$. For Euclidean label distances with $\mu_y=y$, set $L=1$.

\par\endgroup

\begingroup
\subsection{Proof of Theorem~\ref{thm:locus_cover_hardness} (complexity of minimum locus covers)}
\label{app:locus_cover_hardness}
We reduce from Vertex Cover: given a finite simple graph $G=(V,E)$ and a nonnegative integer $k$, decide whether there is $C\subseteq V$ with $|C|\le k$ meeting every edge. This problem is NP-complete~\citep{Karp_1972}.
Write $n=|V|$ and $m=|E|$. We may assume $n\ge1$, since adding an isolated vertex does not change the minimum vertex-cover size.

\paragraph{Construction.}
For each $v\in V$, create a candidate $c_v$ and two private anchors $p_v,q_v$. For each $e\in E$, create a target $t_e$ and a private anchor $a_e$. All these points are distinct. Set
\[
\mathcal Y_G=\{c_v,p_v,q_v:v\in V\}\cup\{t_e,a_e:e\in E\},
\qquad A=\{p_v,q_v:v\in V\}\cup\{a_e:e\in E\}.
\]
Thus $|\mathcal Y_G|=3n+2m$ and $|A|=2n+m$. Define a symmetric distance $d_G$ by setting $d_G(x,x)=0$ and assigning distance one exactly to the unordered pairs
\[
\{c_v,p_v\},\quad\{c_v,q_v\},\quad\{t_e,a_e\},\quad
\{t_e,c_v\}\ \text{when $v$ is an endpoint of $e$}.
\]
Set every other distance between distinct points to two. The triangle inequality holds because each nonzero distance is at most two and the sum of any two nonzero distances is at least two. The cover-size threshold is $B+k$, where $B=2n+m$.

For $S\subseteq\mathcal Y_G$ and weights in the closed probability simplex on $S$, write
\[
F_{\mathbf w}(y)=\sum_{s\in S}w_s d_G(y,s)^2.
\]
The weights may differ for each label, as in Definitions~\ref{def:loc_cover} and~\ref{def:identifying_loc_cover}.

\paragraph{Labels outside the locus.}
If $y\notin S$ has at most one distance-one neighbor in $S$, then $y\notin\Pi(S)$. To see this, fix any probability weights on $S$. If that neighbor $s$ exists and $w_s>0$, then
\[
F_{\mathbf w}(y)=4-3w_s,
\qquad F_{\mathbf w}(s)\le4(1-w_s)<4-3w_s.
\]
Otherwise all positive-weight points have distance two from $y$, so $F_{\mathbf w}(y)=4$. Since the weights sum to one, some $r\in S$ has positive weight, and $F_{\mathbf w}(r)\le4(1-w_r)<4$. In both cases $y$ is not a minimizer. If $S$ is empty, the weights cannot sum to one, so $S$ is not a cover.

Every private anchor has exactly one distance-one neighbor in the entire space. Consequently every ordinary locus cover, and hence every identifying cover, contains $A$.

\paragraph{From a vertex cover to an identifying locus cover.}
Let $C\subseteq V$ be a vertex cover and set
\[
S=A\cup\{c_v:v\in C\}.
\]
Each point in $S$ is the unique minimizer when all weight is placed on that point. For any candidate $c_v$ outside $S$, placing weight $1/2$ on each of $p_v,q_v$ makes $c_v$ the unique minimizer. Indeed, $c_v$ is their only common distance-one neighbor, with objective one. Each support point has objective two, and every other point has objective at least $5/2$.

For each edge $e$, choose an endpoint $v\in C$. Equal weights on $a_e,c_v$ make $t_e$ the unique minimizer: the two support points have distance two from each other and $t_e$ is their only common distance-one neighbor. The same objective comparison applies. Thus $S$ is an identifying locus cover of size $B+|C|$.

\paragraph{From an ordinary locus cover to a vertex cover.}
Let $S$ be any ordinary locus cover. We already know that $A\subseteq S$. Initially choose every $v$ for which $c_v\in S$. If an edge $e=uv$ is still uncovered, neither $c_u$ nor $c_v$ is selected. The only other distance-one neighbor of $t_e$ is $a_e$. The argument above therefore gives $t_e\in S$. For each such uncovered edge, add one endpoint to the chosen vertex set. The resulting set $C$ covers $G$ and satisfies
\[
|C|\le |\{v:c_v\in S\}|+|\{e:t_e\in S\}|
=|S\setminus A|=|S|-B.
\]
Combining the two directions gives
\begin{align*}
G\text{ has a vertex cover of size at most }k
&\iff (\mathcal Y_G,d_G)\text{ has a locus cover of size at most }B+k\\
&\iff (\mathcal Y_G,d_G)\text{ has an identifying locus cover of size at most }B+k.
\end{align*}
If $\tau(G)$ is the minimum vertex-cover size, both types of minimum locus cover have size $B+\tau(G)$.

\paragraph{Polynomial running time.}
Use an explicit vertex list and edge list for $G$, and encode $k$ in binary. Let $L$ denote the input bit length, so $n+m=O(L)$. List the $N=3n+2m$ points and write the $N\times N$ distance matrix using the rules above. Every entry is $0$, $1$, or $2$, requiring only two bits. Label comparisons and a sequential scan of the edge list decide an entry in $O((n+m+1)\log(n+m+2))$ bit operations. The full matrix can therefore be written in $O((n+m+1)^3\log(n+m+2))$ bit operations. Binary addition computes $B+k$ in time polynomial in $L$, without a loop of length $k$. The output has $O(L^2)$ bits, and the construction takes at most $O(L^4)$ bit operations. This gives a polynomial-time reduction to both cover decision problems and proves the NP-hardness part of Theorem~\ref{thm:locus_cover_hardness}.

\paragraph{Polynomial verification and membership in NP.}
Let $N=|\mathcal Y|$ and let $\Lambda\subseteq\mathcal Y$ be a proposed cover. We represent $\Lambda$ by an $N$-bit indicator vector and first check that $|\Lambda|\leq k$. For each $y\in\mathcal Y$, consider the following rational linear program in $N$ weights and a margin $\gamma$:
\begin{align}
\text{maximize}\quad &\gamma \label{eq:locus_verification_lp}\\
\text{subject to}\quad
&w_s\geq0\quad(s\in\mathcal Y),\qquad
\sum_{s\in\mathcal Y}w_s=1,\qquad
w_s=0\quad(s\notin\Lambda),\nonumber\\
&0\leq\gamma\leq1,\qquad
\sum_{s\in\mathcal Y}w_s\bigl(d^2(y,s)-d^2(z,s)\bigr)+\gamma\leq0
\quad(z\in\mathcal Y\setminus\{y\}).\nonumber
\end{align}
For an ordinary locus cover, fix $\gamma=0$ and test feasibility for every $y$. By the definition of the Fr\'echet objective, these inequalities say exactly that $y$ is a minimizer for weights supported on $\Lambda$. Thus all tests succeed if and only if $\Lambda$ is a locus cover.

For an identifying locus cover, check that the linear program in~\ref{eq:locus_verification_lp} is feasible and has a positive optimum for every $y$. A feasible positive margin makes $y$ the unique minimizer. Conversely, if $y$ is the unique minimizer for some weights, every other label has a strictly larger objective. Taking the minimum of these finitely many differences and $1$ gives a positive feasible margin. If $\mathcal Y=\{y\}$, take $\gamma=1$. Thus restricting $\gamma$ to be at most $1$ does not change which labels can be unique minimizers. The feasible region is closed and lies in $[0,1]^{N+1}$, so a nonempty feasible region attains its maximum. This proves that the test is correct, including when the label space has only one point. If $\Lambda$ is empty, the weights cannot sum to one.

To bound the runtime, let $L$ be the number of bits used to represent the rational distance matrix and $k$. Each program has $N+1$ variables and $O(N)$ constraints. Since $w_s\geq0$, we can enforce $w_s=0$ outside $\Lambda$ by adding $w_s\leq0$. Writing the sum-to-one constraint as two inequalities gives at most $3N+4$ rows, including redundant rows. Each nonconstant coefficient is a difference of squared rational distances. For input fractions $p/q$ and $r/s$, it can be represented as
\[
\frac{p^2s^2-r^2q^2}{q^2s^2}.
\]
If the absolute values of the input numerators and denominators are at most $2^L$, the numerator has absolute value at most $2^{4L+1}$ and the denominator at most $2^{4L}$. Thus each coefficient has $O(L)$ bits and can be formed in polynomial time. Together, the $N$ programs contain $O(N^3)$ rational entries, so their total bit length is polynomial in $L$. Exact rational linear programming decides feasibility and computes a finite optimum in time polynomial in the input bit length; the sign of the optimum can therefore be checked exactly~\citep{Adler_1992}.
Consequently, each proposed cover is verifiable in polynomial time. We can also check in polynomial time that the input matrix defines a metric by testing symmetry, zero diagonal entries, positive off-diagonal entries, and the triangle inequalities. Both decision problems belong to NP, and the reductions above make both NP-complete, proving Theorem~\ref{thm:locus_cover_hardness}.

For a fixed $\Lambda$, these tests also compute its locus, or the labels that can be unique minimizers, in polynomial time. Thus we can efficiently compute the locus of a given set even though finding a minimum cover is NP-hard.

\par\endgroup

\begingroup
\subsection{Minimum locus covers for trees}
\begin{theorem}[Minimum locus cover for trees]
\label{thm:min_locus_trees}
Let $T$ be a nonempty finite tree with positive edge lengths and all vertices as labels. Its leaves form the unique minimum locus cover and an identifying locus cover. For a singleton tree, regard its sole vertex as a leaf.
\end{theorem}
\begin{proof}
For a singleton the claim is immediate. Each leaf is uniquely attained by a point-mass weight on itself. Every internal vertex $v$ lies on a path between distinct leaves $l_1,l_2$. Put $a=d(v,l_1)$, $b=d(v,l_2)$ and give these leaves weights $b/(a+b)$ and $a/(a+b)$, respectively. On their path, writing $t=d(y,l_1)$, the objective is
\[
\frac{b t^2+a(a+b-t)^2}{a+b}=(t-a)^2+ab,
\]
uniquely minimized at $v$. For a vertex off the path, projection onto the path strictly decreases both distances, so it cannot minimize. Thus the leaves identify every vertex.

Conversely, let $l$ be a leaf omitted from any proposed support and let $u$ be its neighbor. For every support point $s\ne l$,
$d(l,s)=d(l,u)+d(u,s)>d(u,s)$.
Every probability-weighted squared-distance objective is therefore strictly smaller at $u$ than at $l$. No locus cover can omit $l$, even if it contains internal vertices. All covers contain the leaves, proving minimality and uniqueness.
\end{proof}

\subsection{Tree pairwise decomposability}
\begin{lemma}[Tree pairwise decomposability]
\label{lemma:tree_pairwise}
For a finite tree with positive edge lengths and all vertices as labels, the locus of any nonempty support $\Lambda$ is its spanning subtree and is pairwise decomposable.
\end{lemma}
\begin{proof}
The spanning subtree is the union of paths $\Gamma(a,b)$ for $a,b\in\Lambda$, including singleton paths. Each path vertex is uniquely attained by the two-endpoint weights used in the preceding proof, so the subtree lies in the union of pairwise loci. A vertex outside the subtree has a neighbor in the direction of that subtree; moving to that neighbor strictly decreases its distance to every support point and hence every weighted objective. Thus there are no other minimizers. The reverse inclusion of the pairwise union in the full locus follows by extending each pair's weights by zeros. For empty support both loci are empty by convention.
\end{proof}

\subsection{Correctness of the phylogenetic cover heuristic}
\begin{theorem}[Algorithm~\ref{alg:phylo} correctness]
\label{thm:algphylo}
Algorithm~\ref{alg:phylo}, using the exact general locus test, returns a locus cover for any nonempty finite phylogenetic label space.
\end{theorem}
\begin{proof}
The list contains finitely many ordered pairs of leaves, including diagonal pairs. Every leaf therefore occurs as an endpoint. An early return is a cover by the exact stopping test. If the loop finishes, every leaf has been added and the final return is the trivial cover. This also handles a singleton label space.
\end{proof}

\subsection{Phylogenetic loci need not be pairwise decomposable}
\begin{proposition}[A phylogenetic counterexample]
\label{prop:phylo_nonpairwise}
There is an unweighted tree whose leaf-only label metric has a locus that is not pairwise decomposable.
\end{proposition}
\begin{proof}
Join a hub to three leaves $a,b,c$ by paths of two unit edges each, and to a fourth leaf $t$ by one unit edge. Only these four leaves are labels. Distances between $a,b,c$ are 4, and each has distance 3 from $t$. Equal weights on $\Lambda=\{a,b,c\}$ give objective 9 at $t$ and $32/3$ at each observed leaf. Thus $t$ is the unique minimizer.

For any pair, say $a,b$, weights $(u,1-u)$ give objective 9 at $t$, but objectives $16(1-u)$ and $16u$ at the support points. At least one is at most 8, so $t$ cannot minimize for any pair. Hence $t\in\Pi(\Lambda)$ but $t$ is outside the union of pairwise loci. In particular, nearest-leaf projection of an unrestricted tree mean does not in general solve the leaf-constrained Fr\'echet problem.
\end{proof}

\subsection{Minimum locus covers for rectangular grids}
\begin{theorem}[Minimum locus cover for rectangular grids]
\label{thm:min_loc_grid}
Let $\mathcal Y=\prod_{j=1}^r\{0,\ldots,m_j\}$, {where $r$ is a positive integer and the $m_j$ are nonnegative integers}, have the unit-edge Cartesian grid metric. If $|\mathcal Y|>1$, any pair of opposite corners is a minimum locus cover. The set of all corners is an identifying locus cover. A singleton grid has its sole vertex as its minimum identifying cover.
\end{theorem}
\begin{proof}
Let $A,B$ be opposite corners, $L=d(A,B)>0$, and fix a target $v$. Write $a=d(v,A)$; then $d(v,B)=L-a$. Every grid vertex $y$ satisfies $d(y,A)+d(y,B)=L$. Giving $A,B$ weights $(L-a)/L,a/L$ therefore gives the objective
\[
(d(y,A)-a)^2+a(L-a).
\]
It is minimized at $v$. One support point can attain only itself, so a two-point cover is minimum when the grid has more than one vertex.

To prove identification by all corners, for each opposite-corner pair choose these weights for the same target $v$ and average the resulting objectives. Every term is minimized at $v$. Equality in their sum requires $d(y,A)=d(v,A)$ for every corner $A$. These equations uniquely determine every nonconstant coordinate: subtract the equation for the origin from that for the corner differing from it only in coordinate $j$ to obtain $m_j-2y_j=m_j-2v_j$. Thus $y=v$, proving unique minimization.
\end{proof}

\subsection{Loci of grid pairs}
\begin{lemma}[Locus of grid subspaces]
\label{lemma:locus_grid_subspace}
In a full rectangular Cartesian grid with unit edges, the locus $\Pi(\{a,b\})$ is exactly the {coordinate box} between $a$ and $b$.
\end{lemma}
\begin{proof}
Coordinatewise projection onto this {box} strictly decreases both distances for any point outside it, ruling out such a minimizer. Within the {box} the endpoints are opposite corners, so the preceding theorem attains every vertex. When $a=b$, both the {box} and locus are the singleton $\{a\}$.
\end{proof}

\subsection{Pairwise decomposability in two-dimensional grids}
\begin{lemma}[Two-dimensional grid pairwise decomposability]
\label{lemma:grid_pairwise}
Let $\mathcal Y=\{0,\ldots,m_1\}\times\{0,\ldots,m_2\}$, {where $m_1,m_2$ are nonnegative integers}, with unit-edge grid distance. For every $\Lambda\subseteq\mathcal Y$, its locus is pairwise decomposable.
\end{lemma}
\begin{proof}
The assertion is immediate for at most one support point. Otherwise, suppose $y$ lies outside every support-pair rectangle. For each $s\in\Lambda$, let $S_s$ contain the signed coordinates in which $s$ differs from $y$, chosen from $\{1^+,1^-,2^+,2^-\}$. Each set is nonempty (since $y\notin\Lambda$), and contains at most one sign per coordinate. The rectangle between two endpoints contains $y$ exactly when the endpoints share no strict coordinate sign. Our assumption therefore says that the sets $S_s$ intersect pairwise.

These sets have a common sign. If one is a singleton this is immediate. Otherwise they are edges of the bipartite graph with parts $\{1^+,1^-\}$ and $\{2^+,2^-\}$. A pairwise-intersecting family of edges in a bipartite graph has a common endpoint: if two distinct edges share endpoint $q$, an edge meeting both without $q$ would join their other endpoints in the same part, which is impossible. If all edges coincide the conclusion is immediate as well.

Thus every support lies strictly on one common coordinate side of $y$. Moving one unit from $y$ in that direction stays within the full grid and decreases every support distance by one. It strictly decreases every probability-weighted objective, so $y$ is not in the full locus. Lemma~\ref{lemma:locus_grid_subspace} gives the desired inclusion in the pairwise union. The other inclusion follows by extending pair weights by zeros.
\end{proof}

\paragraph{Failure in three dimensions.}
On $\{-1,0,1\}^3$, put equal weights on
\[
\Lambda=\{(1,1,1),(1,-1,-1),(-1,1,-1),(-1,-1,1)\}.
\]
For every grid point $x$, $d(x,s)=3-s\cdot x$ for these corners, giving
$\frac14\sum_{s\in\Lambda}d^2(x,s)=9+\|x\|_2^2$.
The origin is the unique minimizer. Any two support corners share a coordinate equal to $+1$ or $-1$; moving the origin one unit toward that value strictly decreases both distances. Thus the origin belongs to no pairwise locus. Higher-dimensional grids require the general locus algorithm unless a separate pairwise property is established.

\subsection{The unweighted complete graph}
\begin{theorem}[Trivial locus cover for the complete graph]
\label{thm:trivial_complete}
An unweighted complete graph has no nontrivial locus cover.
\end{theorem}
\begin{proof}
For any nonempty support and probability weights, an omitted label has objective 1. Some support label has positive weight $w$, and its objective is $1-w<1$. Hence no omitted label can minimize. Conversely, every support label is attained by its point-mass weight, so $\Pi(\Lambda)=\Lambda$. Empty support is not a cover of a nonempty label space, and the argument includes the singleton case.
\end{proof}

\subsection{Optimal next-class selection for trees}
\begin{proof}[Proof of Theorem~\ref{thm:active_largest}]
By Lemma~\ref{lemma:tree_pairwise}, $H$ is the spanning subtree of $\Lambda$. For a vertex $v$ outside $H$, uniqueness of tree paths gives a unique first attachment vertex $b(v)$ in $H$. The spanning subtree after adding $v$ is exactly $H\cup\Gamma(v,b(v))$. The new vertices are therefore $\Gamma(v,b(v))\setminus\{b(v)\}$, proving the gain formula. Adding an unobserved vertex already in $H$ gives zero gain; when $H$ is proper, every outside vertex gives positive gain. Maximizing the displayed path count therefore optimizes over all unobserved labels. For unit edges the count equals $d(v,b(v))$. If $H$ is the whole tree, every available addition gives zero gain, as stated.
\end{proof}
\par\endgroup

\section{Algorithms and Time Complexity Analyses}
\label{app:time_complexity}

We provide time complexity analyses for Algorithms~\ref{alg:phylo},~\ref{alg:locus_pairwise}, and~\ref{alg:locus_general}. 

\subsection{Analysis of Algorithm~\ref{alg:phylo} (locus cover for phylogenetic trees)}
We provide Algorithm~\ref{alg:phylo} with comments corresponding to the time complexity of each step.
\begin{algorithm}[]
\caption{Locus cover for phylogenetic trees}\label{alg:phylo}
\begin{algorithmic}
\Require phylogenetic tree $T=(\set V, \set E)$, $\set{Y}=\text{Leaves}(T)$
\State $N \gets |\set{Y}|$
\State $P \gets \text{sortbylength}( [\Gamma(y_i, y_j)]_{i,j \in [N]} )$ \Comment{$N|\set{E}| + N^2\log N$}
\State $P \gets \text{reverse}(P)$ \Comment{$O(N^2)$}
\State $\Lambda \gets \emptyset$
\For{$\Gamma(y_i, y_j)$ in $P$} \Comment{$O(N^2)$}
\If{$\Pi(\Lambda) = \set{Y}$} \Comment{{Algorithm~\ref{alg:locus_general}}}
    \State \Return $\Lambda$ 
\Else
    \State $\Lambda \gets \Lambda \cup \{y_i, y_j\}$
\EndIf
\EndFor
\State {\Return $\Lambda$}
\end{algorithmic}
\end{algorithm}

{First compute the shortest-path distances in the tree. The heuristic considers $O(N^2)$ pairs of endpoints and calls Algorithm~\ref{alg:locus_general} at most once per pair, for a total of $O(N^3)$ linear feasibility problems. Constructing and sorting paths and updating the observed set also take polynomial time. Thus the heuristic runs in polynomial time for rational edge lengths, though it need not return a minimum cover.}

\begingroup
\subsection{Exact computation of a pairwise decomposable locus}
\label{app:pairwise_locus_algorithm}
Suppose the locus is pairwise decomposable and distances are given by a rational matrix. For each pair $a,b\in\Lambda$ and label $y$, define
\[
A_z=d^2(y,a)-d^2(z,a),\qquad B_z=d^2(y,b)-d^2(z,b).
\]
The label $y$ is a Fr\'echet minimizer for weights $(t,1-t)$ exactly when $t\in[0,1]$ and $(A_z-B_z)t\leq-B_z$ for every $z\in\mathcal Y$. Each inequality restricts the allowed values of $t$. Intersecting these constraints with $[0,1]$ gives all weights for which $y$ is a minimizer, including ties.
\begin{algorithm}[ht]
\caption{Exact pairwise locus computation}\label{alg:locus_pairwise}
\begin{algorithmic}

\Require $\Lambda\subseteq\mathcal Y$, rational distance matrix $d$; pairwise decomposability
\State $P\gets\Lambda$
\For{each unordered pair $\{a,b\}\subseteq\Lambda$ with $a\ne b$}
\For{each $y\in\mathcal Y$}
\State $I\gets[0,1]$
\For{each $z\in\mathcal Y$}
\State $A\gets d^2(y,a)-d^2(z,a)$; $B\gets d^2(y,b)-d^2(z,b)$
\State $I\gets I\cap\{t\in\mathbb R:(A-B)t\leq-B\}$
\EndFor
\If{$I\ne\emptyset$}
\State $P\gets P\cup\{y\}$
\EndIf
\EndFor
\EndFor
\State \Return $P$
\end{algorithmic}
\end{algorithm}

The interval test returns precisely $\Pi(\{a,b\})$. Every observed label is a minimizer when all weight is placed on it, so we initialize the result with $\Lambda$. Taking the union of the pairwise loci then gives $\Pi(\Lambda)$ by pairwise decomposability. This also handles observed sets with zero or one label.

After distances have been computed, the algorithm performs $O(K^2N^2+N)$ rational arithmetic operations and comparisons. An interval endpoint is either $0$, $1$, or the ratio $-B_z/(A_z-B_z)$ from a single constraint. Intersecting the constraints only selects the largest lower bound and smallest upper bound. Each endpoint therefore has polynomial bit length, so the algorithm runs in polynomial time even when we account for the cost of rational arithmetic.

\subsection{Polynomial-time computation of a general locus}
\label{app:general_locus_algorithm}
For each label, Algorithm~\ref{alg:locus_general} checks whether some probability weights on $\Lambda$ make that label a minimizer. The distances are fixed, so this is a linear feasibility problem.
\begin{algorithm}[ht]
\caption{Exact general locus computation by linear programming}\label{alg:locus_general}
\begin{algorithmic}

\Require $\Lambda\subseteq\mathcal Y$, explicit rational distance matrix $d$
\State $P\gets\emptyset$
\For{each $y\in\mathcal Y$}
\State Test feasibility in variables $(w_\lambda)_{\lambda\in\Lambda}$ of:
\State \hspace{1em}$w_\lambda\geq0\ (\lambda\in\Lambda),\qquad\sum_{\lambda\in\Lambda}w_\lambda=1$
\State \hspace{1em}$\sum_{\lambda\in\Lambda}w_\lambda[d^2(y,\lambda)-d^2(z,\lambda)]\leq0\quad(z\in\mathcal Y)$
\If{feasible}
\State $P\gets P\cup\{y\}$
\EndIf
\EndFor
\State \Return $P$
\end{algorithmic}
\end{algorithm}

\begin{proposition}[Exact locus computation]
Given a rational distance matrix, Algorithm~\ref{alg:locus_general} returns $\Pi(\Lambda)$ in polynomial time in the input bit length. It uses $N$ LP feasibility checks with $K$ variables and $O(N+K)$ constraints each.
\end{proposition}
\begin{proof}
The constraints require probability weights on $\Lambda$ for which the objective at $y$ is no greater than at any other label. Thus the problem is feasible exactly when $y\in\Pi(\Lambda)$, including when $y$ ties with another minimizer. If $\Lambda=\emptyset$, the weights cannot sum to one, so the algorithm returns the empty locus. These are the constraints from Appendix~\ref{app:locus_cover_hardness} with $\gamma=0$ and the weights outside $\Lambda$ removed. The same coefficient-size argument applies. Each of the $N$ feasibility problems can therefore be solved in polynomial time.
\end{proof}

For a connected graph with positive rational edge lengths, first compute the shortest-path distance matrix in polynomial time and then apply the algorithm. If the rational distance matrix is already available, use it directly. To find the labels that can be unique minimizers, use the positive-margin test from Appendix~\ref{app:locus_cover_hardness} instead. 
\par\endgroup

\section{Additional Experiments}
\label{app:experiments}

\begin{figure}[t!]
    \begin{center}
        \includegraphics[width=.6\columnwidth]{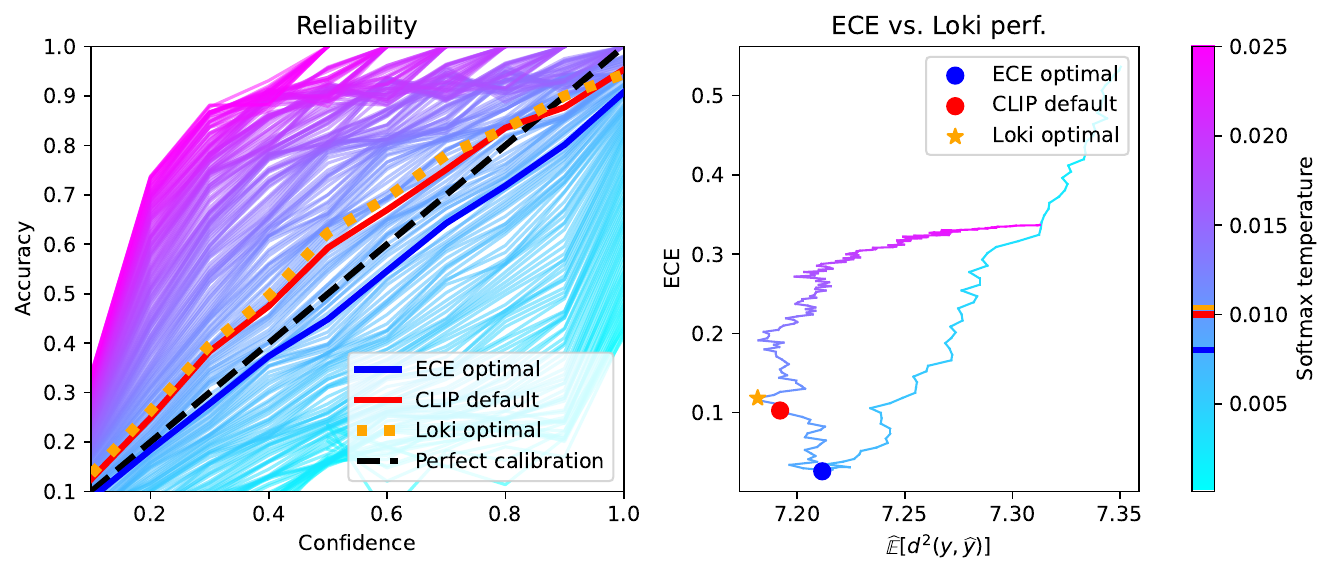}
        \includegraphics[width=.35\columnwidth]{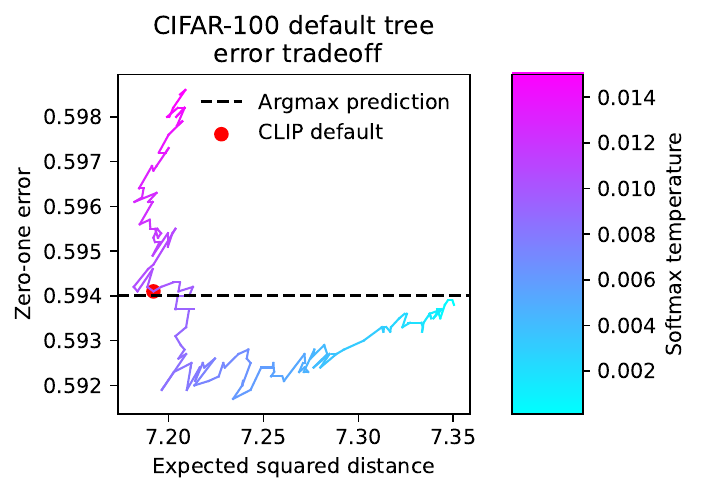} 
        \\
        \includegraphics[width=.6\columnwidth]{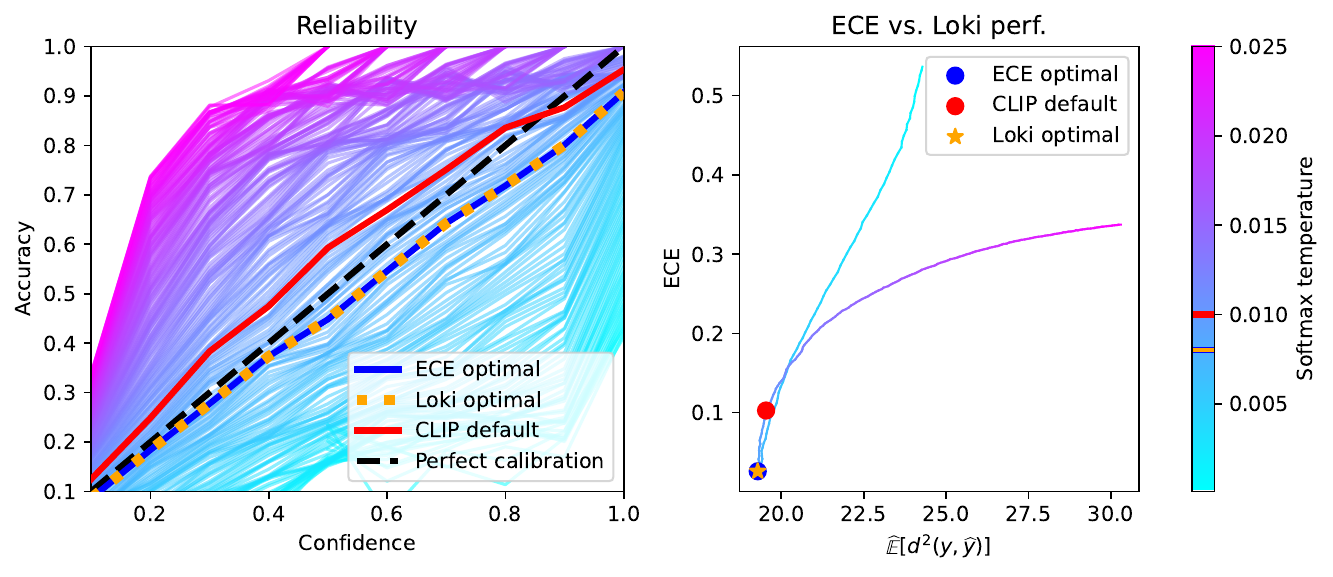}
        \includegraphics[width=.35\columnwidth]{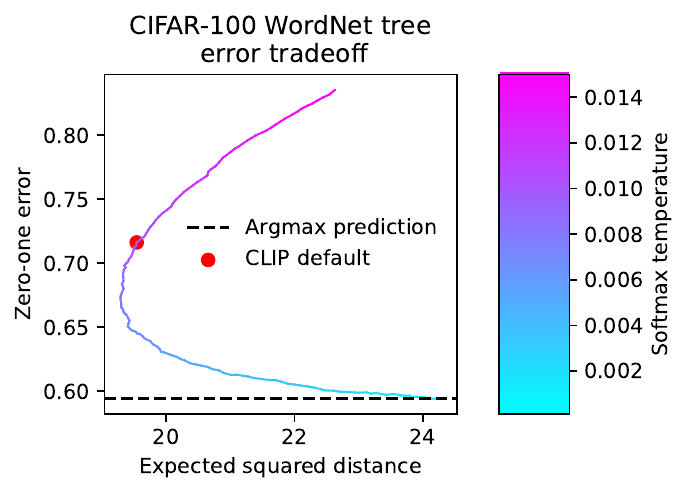} 
    \end{center}
    \caption{
    \textbf{CLIP on CIFAR-100 with the WordNet hierarchy.} 
    (Left) Reliability diagrams across a range of Softmax temperatures, highlighting the CLIP default temperature, the optimal temperature for $\loki$, and the minimizer of the Expected Calibration Error (ECE). All three are well-calibrated. 
    (Center) Tradeoff between optimizing for ECE and the expected squared distance. As with the reliability diagrams, the CLIP default temperature, the $\loki$-optimal temperature, and the ECE-optimal temperature are similar. 
    (Right) Tradeoff between optimizing for zero-one error and the expected squared distance. Depending on the metric space, temperature scaling can improve \textit{accuracy}. 
    }
    \label{fig:app_calibration}
    \vspace{6pt}
\end{figure}

\subsection{Additional calibration experiments}
In the main text, we evaluated the effect of calibrating the Softmax outputs on the performance of $\loki$ using a metric space based on the internal features of the CLIP model.
Here, we perform the same analysis using two external metric spaces.

\paragraph{Setup}
We perform our calibration analysis using the default and WordNet tree. 

\paragraph{Results}
Our results are shown in Figure~\ref{fig:app_calibration}. 
Like our experiment using an internally-derived metric, we find that the optimal Softmax temperature is close to the CLIP default and the optimal temperature for calibration. 
For the default tree, we again found that temperature scaling can be used to improve accuracy using $\loki$. 

\paragraph{Setup}
We calibrate via Softmax temperature scaling \cite{pmlr-v70-guo17a} using CLIP on CIFAR-100. 
We do not use an external metric space, and instead use Euclidean distance applied to the CLIP text encoder. 

\paragraph{Results}
The reliability diagram in Figure~\ref{fig:calibration} shows that the optimal Softmax temperature for $\loki$ is both close to the default temperature used by CLIP and to the optimally-calibrated temperature. 
In Figure~\ref{fig:calibration} (right), we find that appropriate tuning of the temperature parameter \textit{can lead to improved accuracy with CLIP}, even when no external metric space is available.

\subsection{Ablation of $\loki$ formulation}
$\loki$ is based on the Fréchet mean, which is defined as $\argmin_{y \in \mathcal{Y}} \sum_{i=1}^K \mathbf{P}_{\lambda_i|x} d^{2}(y, \lambda_i)$. 
However, this is not the only approach that can be considered. For example, the Fréchet \textit{median}, often used in robust statistics, is defined as $\argmin_{y \in \mathcal{Y}} \sum_{i=1}^K \mathbf{P}_{\lambda_i|x} d(y, \lambda_i)$, without squaring the distances. 
More generally, we define $\argmin_{y \in \mathcal{Y}} \sum_{i=1}^K \mathbf{P}_{\lambda_i|x} d^{\beta}(y, \lambda_i)$ and evaluate different choices of $\beta$. 

\paragraph{Setup} 
We conduct this experiment on ImageNet using SimCLR as our pretrained classifier with $K=$ 250, 500, and 750 randomly selected classes. 

\paragraph{Results} 
{Among the exponents tested, $\beta=2$ gives the lowest mean squared distance for each $K$.}

\begin{table}[h]
\caption{Expected squared distances of SimCLR+Loki alternatives on ImageNet. 
We ablate over the choice of distance exponent in $\loki$ (where $\textcolor{red}{\beta} = 2$, corresponding to the Fréchet mean), including the Fréchet median ($\textcolor{red}{\beta} = 1$). 
That is, we tune
$\textcolor{red}{\beta} : \hat{y} \in \text{arg min}_{y \in \mathcal{Y}} \sum_{i=1}^K \mathbf{P}_{\lambda_i|x} d^{\textcolor{red}{\beta}}(y, \lambda_i)$ {and find that $\textcolor{red}{\beta}=2$ gives the lowest loss for each $K$.} 
}
\begin{center}
    \begin{tabular}{l|l|l|l}
    \toprule
    $\textcolor{red}{\beta}=$                & $K=250$ & $K=500$   & $K=750$ \\
    \midrule
    \rowcolor[gray]{.9} 
    0.5                &    61.28      &   46.57      &    36.31     \\
    1                  &    52.98      &   41.06      &    33.14     \\
    \rowcolor[gray]{.9} 
    {\bf 2 (Loki)}     &{\bf46.78}     &{\bf37.76}    &{\bf29.88}    \\
    4                  &    47.99      &   39.58      &    34.65     \\
    \rowcolor[gray]{.9} 
    8                  &    65.29      &   58.42      &    54.90     \\
    \bottomrule
    \end{tabular}
\end{center}
\end{table}

\subsection{Comparison across metric spaces}
\begingroup
{For a metric with positive diameter $D$, dividing the squared-distance loss by $D^2$ puts it in $[0,1]$. This removes differences in scale, but the metrics still assign different costs to different mistakes. For the unweighted complete graph, the normalized loss is the zero-one loss. Table~\ref{app:cross_metric_table} reports these normalized losses; WordNet gives the lowest value in this comparison.}
\par\endgroup

\begin{table}[h]
\caption{Comparison across metric spaces for CLIP on CIFAR-100 by normalizing by the squared diameter of the metric space: 
$\mathbb{E}[d^2(y, \hat{y})] / \textcolor{red}{\text{diam}(G)^2}$.}
\label{app:cross_metric_table}
\begin{center}
    \begin{tabular}{l|l|l}
    \toprule
    Metric space $G$ & $\textcolor{red}{\text{diam}(G)^2}$ & $\mathbb{E}[d^2(y, \hat{y})] / \textcolor{red}{\text{diam}(G)^2}$ \\
    \midrule
    \rowcolor[gray]{.9} 
    Complete graph   & 1                  & 0.5941 (0-1 error)                               \\
    Default tree     & 16                 & 0.4493                                           \\
    \rowcolor[gray]{.9} 
    {\bf WordNet tree}     & 169                & {\bf 0.1157}                                           \\
    CLIP features    & 0.9726             & 0.2686 \\
    \bottomrule
    \end{tabular}
\end{center}
\end{table}

\section{Experimental Details}
\label{app:experimental_details}
In this section, we provide additional details about our experimental setup.\footnote{Code implementing all of our experiments is available here: \url{https://github.com/SprocketLab/loki}.} 

\subsection{CLIP experiments}
All experiments are carried out using CLIP frozen weights. There are no training and hyperparameters involved in experiments involving CLIP, except for the Softmax temperature in the calibration analysis. 
We evaluted using the default CIFAR-100 test set. 
The label prompt we use is "a photo of a [class\_name]." 

\subsection{ImageNet experiments} 
To construct our datasets, we randomly sample 50 images for each class from ImageNet as our training dataset then use the validation dataset in ImageNet to evaluate $\loki$'s performance.
We extract the image embeddings by using  SimCLRv1\footnote{Embeddings are extracted from the checkpoints stored at: \url{https://github.com/tonylins/simclr-converter}} and train a baseline one-vs-rest classifier.
We use WordNet phylogenetic tree as the metric space. 
The structure of phylogenetic tree can be found at here\footnote{\url{https://github.com/cvjena/semantic-embeddings/tree/master/ILSVRC}}. 
We test different numbers of observed classes, $K$, from 1000 classes.
Observed classes are sampled in two ways, uniformly and Gibbs distribution with the concentration parameter 0.5.

\subsection{LSHTC experiments} 
We generate a summary graph of the LSHTC class graph (resulting in supernodes representing many classes) by iteratively:
\begin{enumerate}
    \item randomly selecting a node or supernode
    \item merging its neighbors into the node to create a supernode
\end{enumerate}
until the graph contains at most 10,000 supernodes.
In the LSHTC dataset, each datapoint is assigned to multiple classes. 
We push each class to its supernode, then apply majority vote to determine the final class. 
We test different numbers of observed classes, $K$, from 10,000 classes. 
We collect datapoints which are in the observed classes.
Then split half of dataset as training dataset and make the rest as testing dataset, including those datapoints which are not in the observed classes. 
We train a baseline classifier using a 5-NN model and compare its performance with $\loki$.

\section{Broader Impacts and Limitations}
\label{app:broader_impacts}
As a simple adaptor of existing classifiers, it is possible for our method to inherit biases and failure modes that might be present in existing pretrained models. 
On the other hand, if the possibility of harmful mis-classifications are known a priori, information to mitigate these harms can be baked into the metric space used by $\loki$. 
Finally, while we have found that $\loki$ often works well in practice, it is possible for the per-class probabilities that are output by a pretrained model to be sufficiently mis-specified relative to the metric over the label space, or for the metric itself to be mis-specified. 
Nonetheless, we have found that off-the-shelf or self-derived metric spaces to work well in practice.



\section{Random Locus Visualizations}
\label{app:vis}
\begin{figure}[h]
\centerline{
\includegraphics[width=\columnwidth]{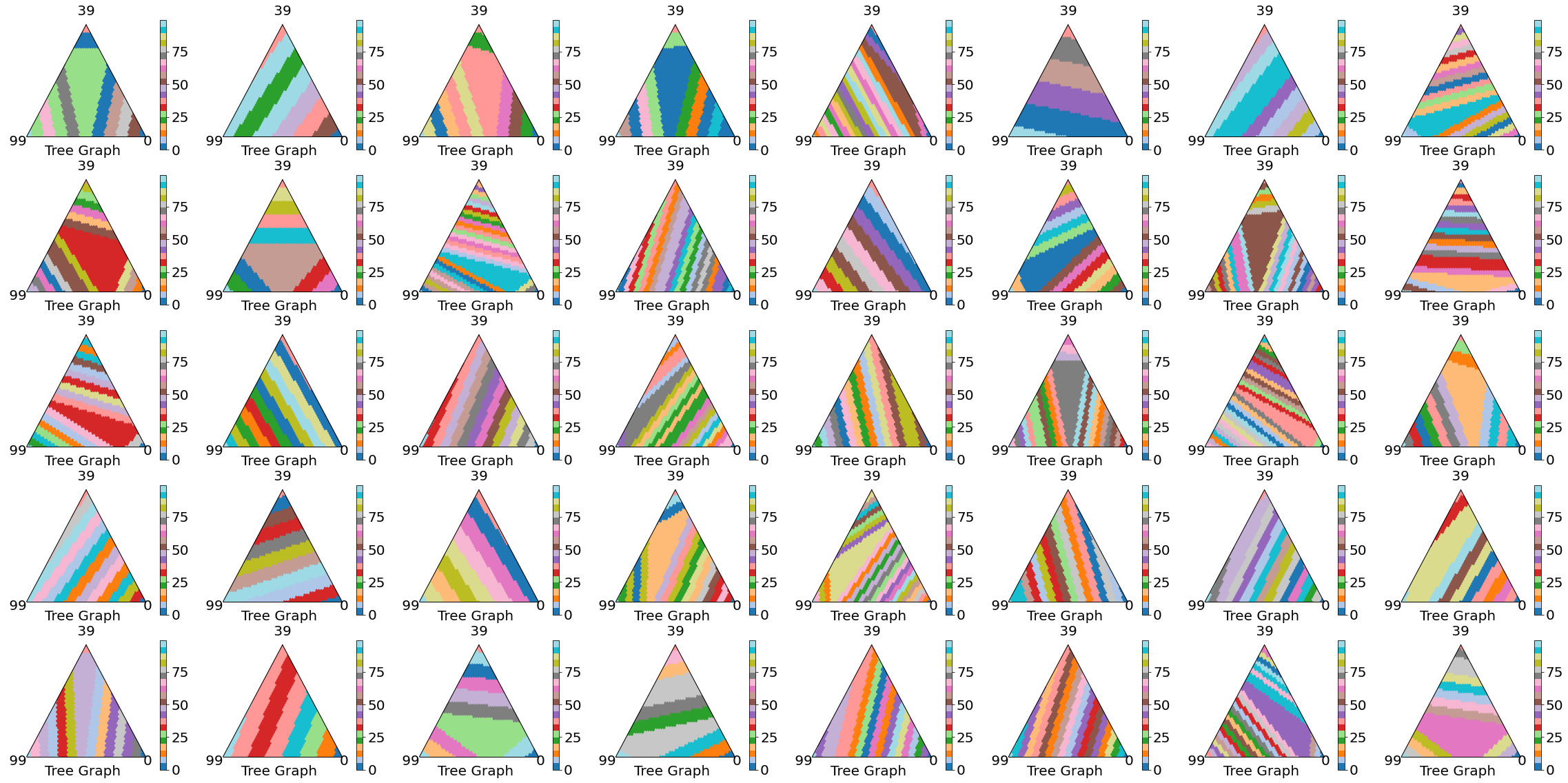}
}
\vspace{-.1in}
\caption{Classification regions in the probability simplex of 3-class classifiers faced with a 100-class problem where the classes are related by random trees graphs.}
\label{fig:tree_loci}
\end{figure}

\begin{figure}[h]
\centerline{
\includegraphics[width=\columnwidth]{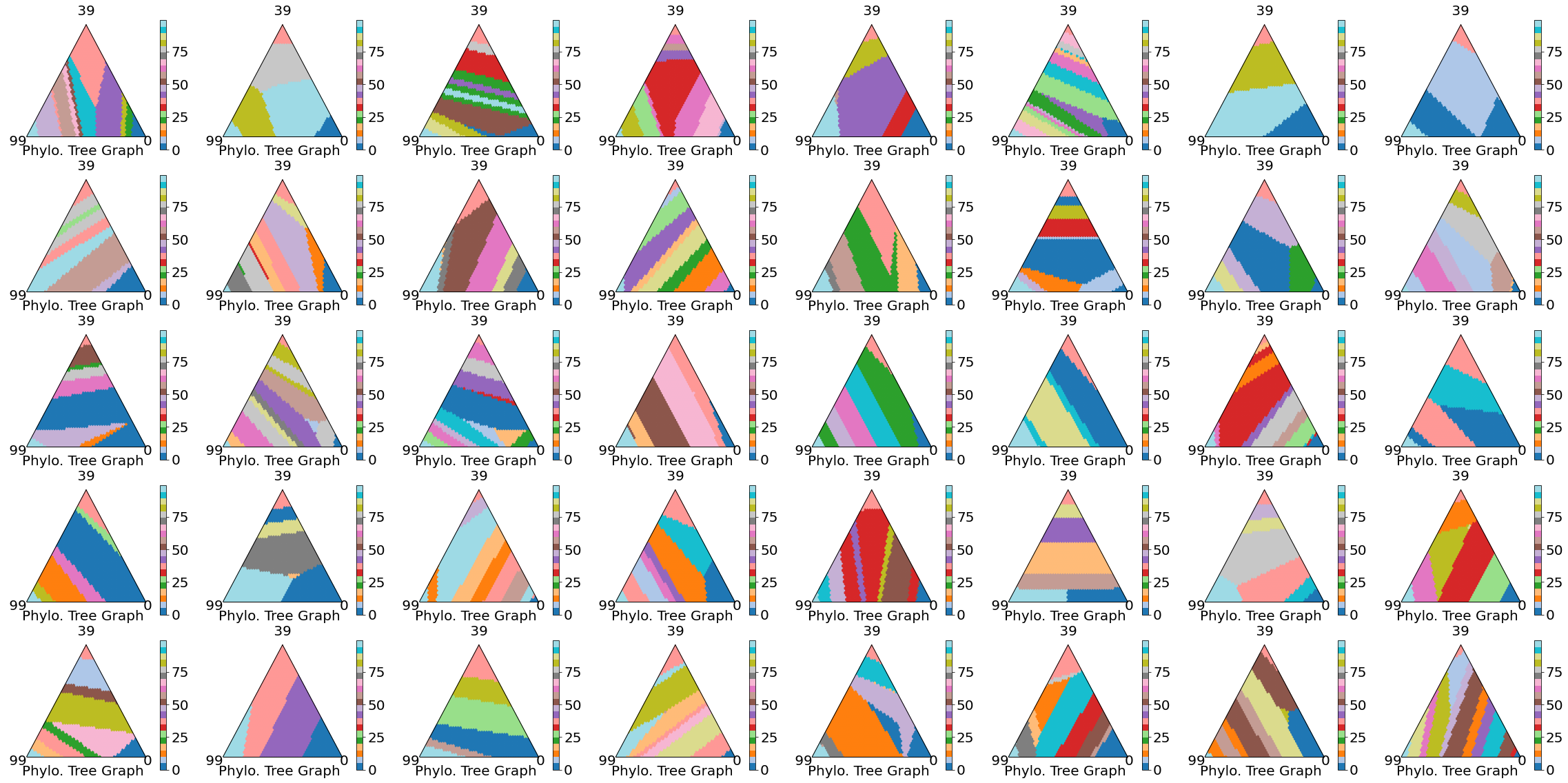}
}
\vspace{-.1in}
\caption{Classification regions in the probability simplex of 3-class classifiers faced with a 100-class problem where the classes are related by random phylogenetic trees graphs.}
\label{fig:phylotree_loci}
\end{figure}

\begin{figure}[h]
\centerline{
\includegraphics[width=\columnwidth]{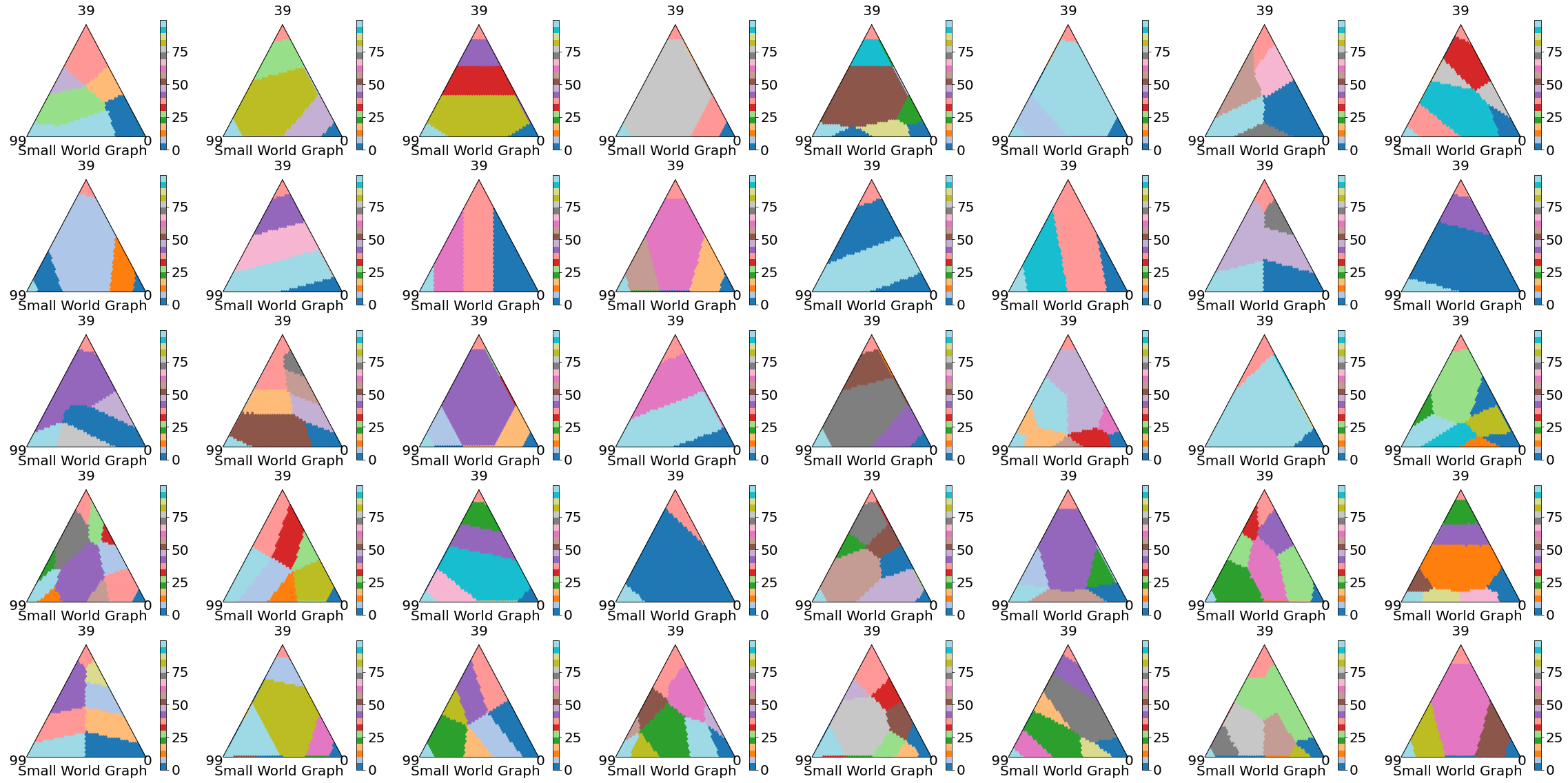}
}
\vspace{-.1in}
\caption{Classification regions in the probability simplex of 3-class classifiers faced with a 100-class problem where the classes are related by random small-world graphs generated by a Watts–Strogatz model.}
\label{fig:smallworld_loci}
\end{figure}

\begin{figure}[h]
\centerline{
\includegraphics[width=\columnwidth]{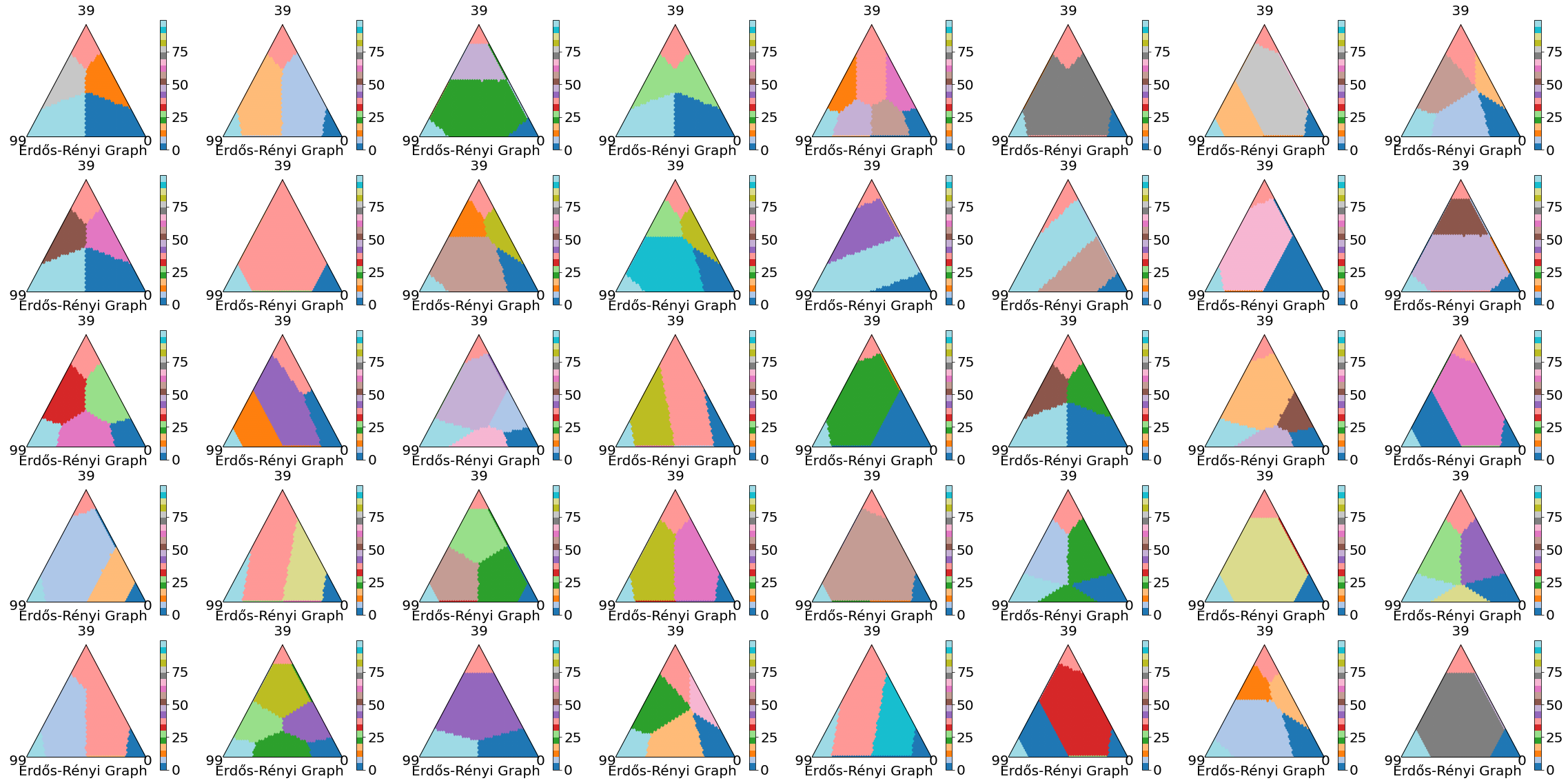}
}
\vspace{-.1in}
\caption{Classification regions in the probability simplex of 3-class classifiers faced with a 100-class problem where the classes are related by random Erd\H{o}s-Rényi graphs.}
\label{fig:erdos_loci}
\end{figure}

\begin{figure}[h]
\centerline{
\includegraphics[width=\columnwidth]{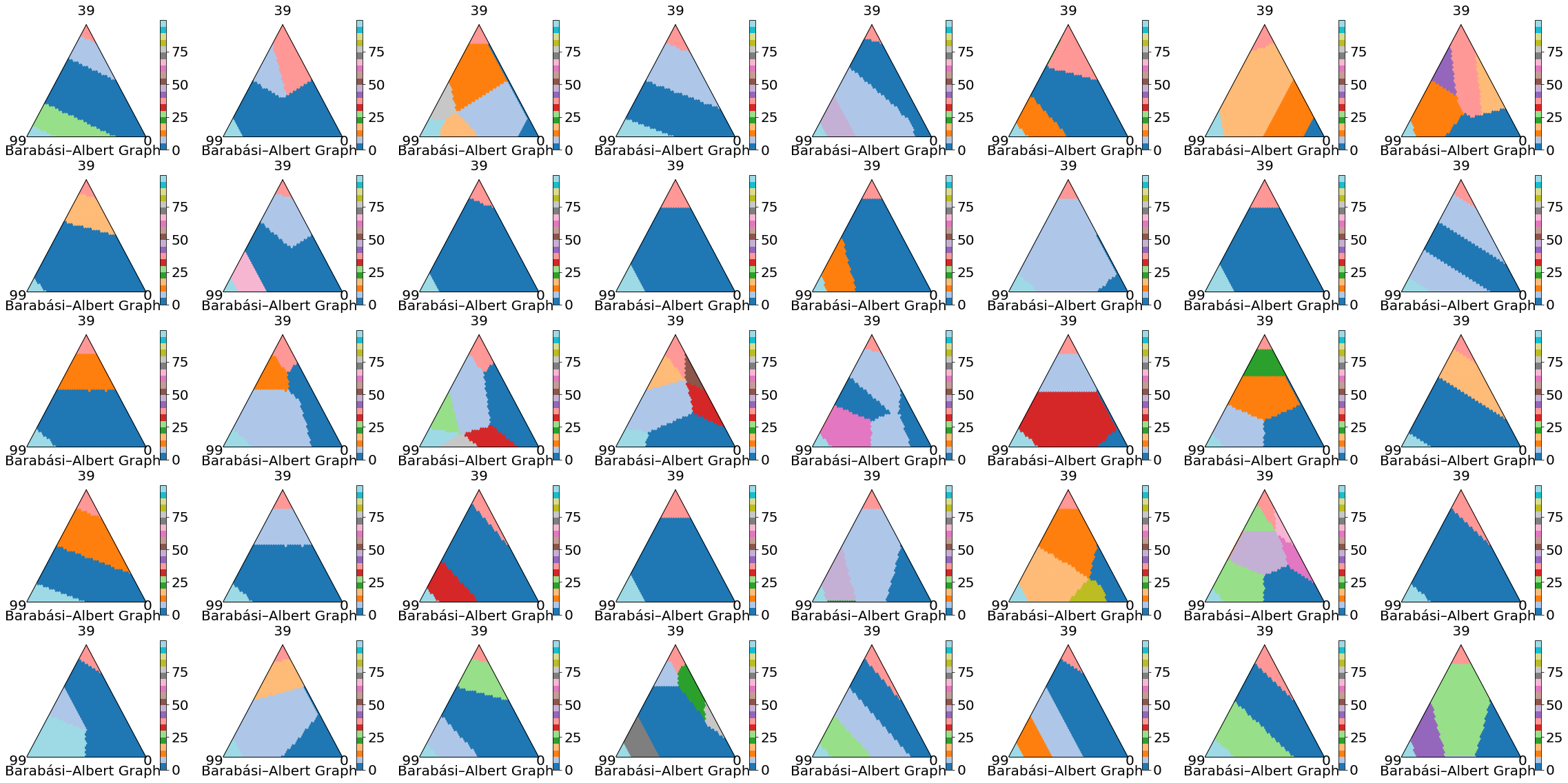}
}
\vspace{-.1in}
\caption{Classification regions in the probability simplex of 3-class classifiers faced with a 100-class problem where the classes are related by random Barabási–Albert graphs.}
\label{fig:albert_loci}
\end{figure}

In Figures~\ref{fig:banner},~\ref{fig:tree_loci},~\ref{fig:phylotree_loci},~\ref{fig:smallworld_loci},~\ref{fig:erdos_loci}, and~\ref{fig:albert_loci}, we provide visualizations of classification regions of the probability simplex when using $\loki$ with only three observed classes out of 100 total classes, and different types of random metric spaces. 
The first example in Figure~\ref{fig:banner} shows the prediction regions on the probability simplex when using standard $\argmax$ prediction---the three regions correspond to predicting one of the three classes (0, 39, and 99) and no regions corresponding to any of the other classes $\{1, ..., 38, 40, ..., 98\}$. 

On the other hand, the other three examples in Figure~\ref{fig:banner} and Figures~\ref{fig:tree_loci},~\ref{fig:phylotree_loci},~\ref{fig:smallworld_loci},~\ref{fig:erdos_loci}, and~\ref{fig:albert_loci} all show the prediction regions on the probability simplex when using $\loki$. 
Figure~\ref{fig:tree_loci} shows this for random tree graphs.
Here, the prediction regions are striped or contain a single triangle-shaped region in the center---these correspond, respectively, to intermediate classes along branches of the tree leading up from the observed class and the prediction region formed by the common-most parent node. 
Figure~\ref{fig:phylotree_loci} shows similar regions, although these are more complex and are thus more difficult to interpret, as phylogenetic trees are metric subspaces equipped with the induced metric from trees. 
Furthermore, in order to generate phylogenetic trees with 100 leaves, we needed to create much larger trees than the ones used for Figure~\ref{fig:tree_loci}, which led to narrower prediction regions due to the higher graph diameter. 
Finally, Figures~\ref{fig:smallworld_loci},~\ref{fig:erdos_loci}, and~\ref{fig:albert_loci} each show the prediction regions using $\loki$ when the metric space is a random graph produced using Watts-Strogatz, Erd\H{o}s-Rényi, and Barabási–Albert, models respectively. 
These prediction regions are more complex and represent complex relationships between the classes.

\vfill

%
%
%
%
%

\end{document}